\documentclass[11pt]{article}
\usepackage{fullpage}
\usepackage[margin=1in]{geometry}

\usepackage{graphicx}
\usepackage{natbib}
\usepackage{xcolor}
\usepackage{hyperref}
\usepackage{algorithm,algpseudocode}
\usepackage{algorithmicx}
\usepackage{multicol}
\usepackage{multirow}
\usepackage{hhline}
\usepackage{amsmath}
\usepackage{mathtools}
\usepackage[capitalize,noabbrev]{cleveref}
\usepackage{float}
\usepackage{array}
\usepackage{xspace}
\usepackage{subfigure}
\newcolumntype{L}[1]{>{\raggedright\let\newline\\\arraybackslash\hspace{0pt}}m{#1}}
\newcolumntype{C}[1]{>{\centering\let\newline\\\arraybackslash\hspace{0pt}}m{#1}}
\newcolumntype{R}[1]{>{\raggedleft\let\newline\\\arraybackslash\hspace{0pt}}m{#1}}

\RequirePackage[T1]{fontenc} 
\RequirePackage[tt=false, type1=true]{libertine} 
\RequirePackage[varqu]{zi4} 
\RequirePackage[libertine]{newtxmath}

\usepackage{nicefrac}

\newtheorem{dfn}{\textbf{Definition}}

\newtheorem{lem}{\textbf{Lemma}}
\newtheorem{thm}{\textbf{Theorem}}
\newtheorem{pro}{\textbf{Proposition}}
\newtheorem{cor}{\textbf{Corollary}}
\newcommand{\cali}{{\mathcal{I}}}

\newcommand{\cald}{{\mathcal{D}}}

\newcommand{\calq}{{\mathcal{Q}}}
\newcommand{\calh}{{\mathcal{H}}}

\newcommand{\caly}{{\cal Y}}

\newcommand{\calc}{{\mathcal{C}}}

\newcommand{\bbz}{{\mathbb{Z}}}

\newcommand{\by}{\boldsymbol{y}}

\newcommand{\bu}{\boldsymbol{u}}
\newcommand{\bq}{\boldsymbol{q}}
\newcommand{\bw}{\boldsymbol{w}}

\newcommand{\ex}{\mathbb{E}}
\newcommand{\pr}{\mathbb{P}}

\newcommand{\opt}{\texttt{OPT}\xspace}
\newcommand{\alg}{\texttt{ALG}\xspace}
\newcommand{\CR}{\texttt{CR}\xspace}
\newcommand{\drcr}{\texttt{DRCR}\xspace}

\newcommand{\dsrppp}{\texttt{DSR}\xspace}

\newcommand{\rsrpip}{\texttt{RSR}\xspace}

\newcommand{\pip}{\texttt{PIP}\xspace}
\newcommand{\ppp}{\texttt{PPP}\xspace}

\newcommand{\rhs}{\texttt{RHS}\xspace}
\newcommand{\lhs}{\texttt{LHS}\xspace}

\newcommand{\crsrppp}{\chi\xspace}
\newcommand{\crsrpip}{\zeta\xspace}

\newcommand{\pfa}{\texttt{PFA}\xspace}

\DeclareMathOperator*{\argmax}{arg\,max}
\DeclareMathOperator*{\argmin}{arg\,min}

\begin{document}

\title{Online Algorithms with Uncertainty-Quantified Predictions}

\author{Bo~Sun\thanks{University of Waterloo. Email: {\tt bo.sun@uwaterloo.ca}.}
\and
Jerry~Huang\thanks{California Institute of Technology. Email: {\tt jyhuang@caltech.edu}.}
\and
Nicolas~Christianson\thanks{California Institute of Technology. Email: {\tt nchristianson@caltech.edu}.}
\and
Mohammad~Hajiesmaili\thanks{University of Massachusetts Amherst. Email: {\tt hajiesmaili@cs.umass.edu}.} 
\and 
Adam~Wierman\thanks{California Institute of Technology. Email: {\tt adamw@caltech.edu}.}
\and
Raouf Boutaba\thanks{University of Waterloo. Email: {\tt rboutaba@uwaterloo.ca}.}
}

\begin{titlepage}
\maketitle

\thispagestyle{empty}

\begin{abstract}
The burgeoning field of algorithms with predictions studies the problem of using possibly imperfect machine learning predictions to improve online algorithm performance. While nearly all existing algorithms in this framework make no assumptions on prediction quality, a number of methods providing \emph{uncertainty quantification (UQ)} on machine learning models have been developed in recent years, which could enable additional information about prediction quality at decision time. In this work, we investigate the problem of optimally utilizing uncertainty-quantified predictions in the design of online algorithms. In particular, we study two classic online problems, ski rental and online search, where the decision-maker is provided predictions augmented with UQ describing the likelihood of the ground truth falling within a particular range of values. We demonstrate that non-trivial modifications to algorithm design are needed to fully leverage the UQ predictions. 
Moreover, we consider how to utilize more general forms of UQ, proposing an online learning framework that learns to exploit UQ to make decisions in multi-instance settings.
\end{abstract}
\end{titlepage}

\maketitle
\section{Introduction}
Classic online algorithms are designed to ensure worst-case performance guarantees. However, such algorithms are often overly pessimistic and perform poorly in real-world applications since worst-case instances rarely occur.
To address the pessimism of these algorithms, a recent surge of work has investigated the design of algorithms utilizing machine-learned predictions~\cite{mitzenmacher2020algorithms,lykouris2021competitive,purohit2018improving}. In this line of research, an algorithm is given additional information on the problem instance in the form of predictions or ``advice'', possibly from a machine learning model. 
Notably, it is typically the case that no assumptions are made on predictions' quality. Thus, algorithms must treat them as ``untrusted'', seeking to exploit them when they are accurate while ensuring worst-case guarantees when they are not. 

Driven by safety-critical applications, 
uncertainty quantification (UQ) has recently become a prominent field of research in machine learning. UQ aims to provide quantitative measurements on machine learning models' uncertainty about their predictions.
One of the state-of-the-art methods for UQ is \textit{conformal inference}~\cite{10.5555/645528.657641,10.5555/1062391,Papadopoulos2002InductiveCM}, which can transform the predictions of any black-box algorithm into a prediction interval (or prediction set) that contains the true value with high probability. 
Although UQ has been widely used for general decision-making under uncertainty, as in~\cite{vovk2018conformal,marusich2023using,sun2023maximum}, there has been limited study on its use for online problems.
Thus, the key question we aim to answer in this paper is: 
\begin{center}
\textit{How can we incorporate uncertainty-quantified predictions into the design of competitive online algorithms?}
\end{center}
To address the above question, we require a new design objective that interpolates between worst-case analysis and average-case analysis for online algorithms. 
Algorithms augmented with UQ predictions have access to both predictions of future inputs and the associated prediction quality, which can be leveraged to improve upon worst-case performance; however, UQ predictions often cannot exactly reconstruct distributional information to enable typical average-case guarantees.
As such, we design algorithms to minimize a new performance metric, which we term \textit{distributionally-robust competitive ratio} (\drcr): we seek algorithms that perform well on the worst-case distribution drawn from the ambiguity set determined by a given UQ. 

In particular, this paper makes contributions in threefold for designing online algorithms that leverage UQ predictions.

\paragraph{Optimal online algorithms under distributionally-robust analysis}

We frame the online algorithms with UQ predictions as a distributionally-robust online algorithm design problem.
Then we design online algorithms using probabilistic interval predictions for the ski rental and online search problems, in both cases showing that they attain the optimal \drcr (see Theorems~\ref{thm:sr-rand} and~\ref{thm:os-general}). These two problems have played a key role in the development of learning-augmented algorithms and thus are natural problems with which to begin the study of UQ for online algorithms.

\paragraph{Optimization-based algorithmic approach} 
Technically, we propose an optimization-based approach for incorporating UQ predictions into online algorithms.
The approach consists of building an ancillary optimization problem based on predictions with the objective of minimizing \drcr for hard instances of the problem.
The solution of the optimization problem then yields the optimal algorithm design, considering the provided UQ.
This approach is general in the sense that the optimization can be tuned based on the specific forms of the predictions and design goals, and, thus, this approach can potentially be applied to incorporate other forms of UQ predictions and to devise algorithms for other online problems beyond ski rental and online search.

\paragraph{Online learning for exploiting UQs across multiple instances} 
Finally, we propose an online learning approach that learns to exploit general forms of UQ across multiple problem instances. Here, the probabilistic interval predictions may be imperfect (e.g., due to non-exchangeability of the data) or alternative notions of UQ are employed. We show that, under mild Lipschitzness conditions, one can obtain sublinear regret guarantees with respect to solving the full optimization formulation of the \drcr problem. {We demonstrate the regret guarantees obtained by this framework in the ski rental and online search problems.} Moreover, when problem instances are not fully adversarial (i.e., the distribution generating problem instances is not the worst case for the given UQ), our online learning approach outperforms the optimization-based approach, as we demonstrate in experiments in Section~\ref{sec:learning-UQ}.

\subsection{Related Literature}

\paragraph{Algorithms with untrusted predictions}
A significant body of work has emerged considering the design of algorithms that incorporate untrusted predictions of either problem parameters or optimal decisions \cite{lykouris2021competitive,purohit2018improving, mahdianOnlineOptimizationUncertain2012,mitzenmacher2022algorithms,wei2020optimal,antoniadis2020online,christianson2023optimal,sun2021pareto}. However, in nearly all of these works, predictions are assumed to be point predictions of problem parameters or decisions, i.e., individual untrusted decisions with no further assumptions on quality, uncertainty, probability of correctness, etc. Several recent studies have considered alternative prediction paradigms, including the setting of learning predictions from samples or distributional advice~\cite{anandCustomizingMLPredictions2020,diakonikolas2021learning, besbesIIDDatadrivenDecisionmaking2022, khodak2022learning}, and predictions which are assumed to be correct with a certain, known or unknown probability~\cite{gupta2022augmenting}. Our work is distinguished from these prior results in that we consider a more general class of \textit{uncertainty-quantified} predictions. In particular, our model of probabilistic interval predictions allows for predictions that fall into a certain interval with a given probability, thus generalizing the prediction paradigm of~\cite{gupta2022augmenting} to one more closely matched to uncertainty quantification methods in the machine learning literature.

{
\paragraph{Online learning} 
Our online learning-based approach to utilizing UQ builds on techniques from the online learning literature and, specifically, online learning with side information and data-driven algorithm design. The problem of exploiting additional side information to improve performance in online learning has been widely studied in both bandit \cite{agrawal2014thompson,
% lale2019stochastic,Li_2010, xuOnlineLearningSide2017, wanigasekara2019nonparametric,guan2018nonparametric,rigollet2010nonparametric,10.1214/aos/1015362186, 
slivkins2014contextual,10.1287/opre.2019.1902} and partial/full-feedback \cite{10.1007/978-3-540-72927-3_36, dekelOnlineLearningHint2017,
kuzborskij2020locallyadaptive} settings. Our work is most closely related to the results in \cite{10.1007/978-3-540-72927-3_36}; however, our results go beyond the Lipschitz assumptions on the policy class employed in \cite{10.1007/978-3-540-72927-3_36}, and we show that we can exploit Lipschitzness of \textit{any} problem instance cost upper bound to enable competing against general policies when exploiting UQ in online problems. 
Our online learning formulation is also aligned with the data-driven algorithm design framework~\cite{balcan2020data,balcan2018dispersion} that adaptively selects the parameterized algorithms across multiple instances without using side information.
Our work extends the problem setting by exploring how to utilize the additional UQ predictions in the algorithm selection, and showing regret guarantees under mild Lipschitzness assumptions. 
}

\section{Online Algorithms with UQ Predictions}
\label{sec:alg-UQ}

For an online cost minimization problem, let $\cali$ denote the set of all instances. For each instance $I\in\cali$, let $\alg(A,I)$ and $\opt(I)$ denote, respectively, the (expected) cost attained by an online algorithm $A$ and the cost of the offline solution.
Under the classic competitive analysis framework~\cite{borodin2005online}, online algorithms have \textit{no prior knowledge} of the instance $I$. Algorithmic design is framed as a single-instance min-max problem, with the objective of finding an online algorithm $A$ to minimize the worst-case competitive ratio
\footnote{In an online profit maximization problem, the competitive ratio is defined as the worst-case ratio between the offline optimum and the profit of the online algorithm.}
, i.e., $\max_{I\in\cali} \frac{\alg(A,I)}{\opt(I)}$.

To improve the performance of online algorithms and go beyond worst-case analysis, there has recently been research emerging on algorithms with (untrusted) predictions~\cite{mitzenmacher2022algorithms,lykouris2021competitive,purohit2018improving}.
In an abstract setup, we consider the input instance of an online problem that can be characterized by a critical value $V$ (e.g., the number of skiing days for the ski-rental problem). Machine learning tools can be leveraged to make a prediction $P$ about the critical value $V$. 
In most scenarios, the \textit{quality of the prediction is unknown} to the online decision-maker; hence, the goal of algorithm design with predictions is to guarantee good performance when the prediction is accurate (i.e., consistency) while still maintaining worst-case guarantees regardless of the prediction accuracy (i.e., robustness).  
Let $\cali_P \subseteq \cali$ denote a \textit{consistent} set that contains all instances confirming with the prediction $P$. Then the consistency $\eta$ and robustness $\gamma$ of an online algorithm $A$ are defined as 
\begin{align}\label{eq:dfn-rc}
    \eta = \max_{I \in \cali_{P}} \frac{\alg(A,I)}{\opt(I)}\ \text{and}\  \gamma = \max_{I \in \cali} \frac{\alg(A,I)}{\opt(I)},
\end{align}
which are the worst-case ratios over $\cali_{P}$ and $\cali$, respectively.
Prior work has shown that there exist strong trade-offs between consistency and robustness bounds~\cite{purohit2018improving,wei2020optimal,Balseiro2023,sun2021pareto,bamas2020primaldual}. Therefore, algorithms with predictions usually provide a parameterized class of online algorithms (using a hyper-parameter $\lambda$) to achieve different trade-offs. 
Due to the lack of prediction quality, the selection of the hyper-parameter is left to end users. 

In practice, we often have access to some forms of uncertainty quantification about the prediction of the input instance.
We model an uncertainty-quantified (UQ) prediction by a vector $\theta := \{P; Q\}$, where $P$ is the prediction and $Q$ specifies the quality of the prediction. 
For a given $\theta$, we assume that instance $I$ belongs to a fixed unknown distribution $\xi_\theta$.
If $\xi_\theta$ can be completely specified by $\theta$, the average-case analysis aims to design the online algorithm that can minimize the expected competitive ratio, i.e., $\ex_{\xi_\theta} [\frac{\alg(A,I)}{\opt(I)}]$.
However, in most cases, UQ can only partially specify the instance distribution, and thus an interpolation between the worst-case analysis and average-case analysis is desired.

\subsection{Distributionally-Robust Competitive Analysis}

When UQ can coarsely characterize the instance distribution, for a given $\theta$, 
we can construct an ambiguity set $\cald_\theta$ that includes all instance distributions that conform with UQ. 
An important example of such UQ is probabilistic quantification of prediction correctness, and the ambiguity set contains all distributions that conform with such predictions. 
In this case, an online algorithm $A$ can be designed to minimize the distributionally-robust competitive ratio (\drcr)
\begin{align}
\label{eq:drcr}
\drcr_\theta (A)=\max_{\xi_{\theta}\in\cald_\theta} \ex_{\xi_{\theta}}\left[\frac{\alg(A, I)}{\opt(I)} \right],
\end{align}
which is the worst expected competitive ratio over instance distributions from $\cald_\theta$.
Such an algorithmic design can be considered as an interpolation between worst-case analysis and average-case analysis. 

For average-case analysis of competitive algorithms, the performance can be evaluated by expectation of ratios or ratio of expectations.   
We choose to define the \drcr as the expectation of ratios for two reasons. First, this metric is more commonly considered as the average-case performance measure for the ski rental problem and online search problems (e.g.,~\cite{fujiwara2005average} and~\cite{fujiwara2011average}), which are the focus of this paper. Second, the \drcr defined based on the expectation of ratios can be shown to be a convex combination of the consistency and robustness of the online algorithms with untrusted algorithms.
This connection makes the \drcr more appealing as an extension of the consistency-robustness metric given additional information on the quality of prediction.

\paragraph{Probabilistic interval predictions}
One important class of UQs that can be leveraged for distributionally-robust analysis is \textit{probabilistic interval predictions} (\pip).

\begin{dfn}
% [Probabilistic interval prediction]
\label{dfn:pip}
For $\ell\le u$ and $\delta\in[0,1]$, 
% $\emph{\pip}(\theta) := \emph{\pip}(\ell,u;\delta)$
$\pip(\theta)$ with $\theta = \{\ell,u;\delta\}$
is called a probabilistic interval prediction for a critical value $V$ of an input instance,
% offline \red{parameter} \mo{rename it to input or prediction value? } 
and it predicts that with at least probability $1-\delta$, the true value $V$ is within $[\ell, u]$, i.e., $\pr(V \in [\ell, u]) \ge 1 - \delta.$
\end{dfn}
In the literature of algorithms with untrusted predictions~\cite{mitzenmacher2022algorithms,lykouris2021competitive,purohit2018improving}, 
the untrusted prediction $P$ is a special case of $\pip(\ell,u;\delta)$ when the prediction is a point prediction $\ell = u = P$ and there is no guarantee on this prediction $\delta = 1$. 
% We use $\ppp(P;0)$ to denote the untrusted prediction.

\pip can be obtained through \textit{conformal predictions} \cite{10.5555/645528.657641,10.5555/1062391,Papadopoulos2002InductiveCM}. Given exchangeable data, conformal prediction can transform the outputs of any black-box predictors into a prediction set/interval that can cover the true value with high probabilities.  
In particular, conformal inference certifies that the prediction $P$ over the critical value $V$ is accurate within an error $\varepsilon$ with at least probability $1-\delta$, i.e., $\mathbb{P}(|V-P|\le \varepsilon) \ge 1 - \delta$. 
In this case, the prediction quality is characterized by the prediction error $\varepsilon$ and prediction confidence $\delta$. Equivalently, we can frame this UQ as  a $\pip(\theta)$ over the instance, i.e., $\mathbb{P}(V \in [\ell, u]) \ge 1 - \delta$, where $\ell := P - \varepsilon$ and $u := P + \varepsilon$. 

For a given $\pip$, let $\cali_{\ell,u} \subseteq \cali$ denote a consistent set that contains all instances that confirm with the interval prediction. 
Then the ambiguity set $\cald_\theta$ can include all instance distributions such that $\pr_{\xi_\theta}(I \in \cali_{\ell,u}) \ge 1 -\delta, \forall \xi_\theta\in\cald_\theta$, i.e., under distribution $\xi_\theta$, the probability that an instance $I$ belongs to a set $\cali_{\ell,u}$ is at least $1-\delta$.
Further, we can observe that the worst instance distribution that maximizes the $\drcr$ in Equation~\eqref{eq:drcr} is a two-point distribution, with probability $1-\delta$ for instance $I^\eta$ and probability $\delta$ for instance $I^\gamma$, where $I^\eta = \argmax_{I\in\cali_{\ell,u}} \frac{\alg(A, I)}{\opt(I)}$ and $I^\gamma = \argmax_{I\in\cali} \frac{\alg(A, I)}{\opt(I)}$.
Thus, the \drcr of online algorithm with $\pip$ $\theta$ can be transformed into 
\begin{align}
\label{eq:drcr-trans}
    % &\drcr_\theta (A) =\\ 
    % \max_{\xi\in\cald} \ex_{I\sim \xi}\left[\frac{\alg(A, I)}{\opt(I)} \right] = 
    &(1-\delta) \cdot \max_{I \in \cali_{\ell,u}} \frac{\alg(A, I)}{\opt(I)} + \delta \cdot \max_{I \in \cali} \frac{\alg(A, I)}{\opt(I)} \nonumber\\
    &:= (1-\delta) \cdot \eta + \delta \cdot \gamma, 
\end{align}
where $\eta$ and $\gamma$ are the consistency and robustness of algorithms with untrusted interval predictions. 

{
\subsection{An Optimization-Based Algorithmic Approach}
\label{sec:approach}

{We introduce an optimization-based algorithmic approach for the \textit{single-instance} distributionally-robust analysis that can be leveraged to systematically design online algorithms with UQ predictions.}
We focus on a class of parameterized online algorithms. Let $A(\bw)$ denote the online algorithm with parameter $\bw\in \Omega$, where $\Omega$ is the parameter set. 
The design of an online algorithm augmented by a UQ prediction $\pip(\theta)$ is to find a policy $\pi \in \Pi: {\Theta} \to \Omega$ that maps from $\theta$ to an online algorithm $A(\bw)$. We propose a general optimization-based approach to design the policy by solving an ancillary optimization problem.

We start by constructing a family of representative hard instances $\calh \subseteq \cali$ and parameterized algorithms $\{A(\bw)\}_{\bw\in\Omega}$ for the online problem based on the problem-specific knowledge. 
Let $\alg(\bw,I)$ and $\opt(I)$ denote the costs of online algorithm $A(\bw)$ and offline algorithm under the instance $I\in\calh$. 
Given the prediction $\theta$, we can further determine a subset $\calh_{\ell,u}$ of $\calh$, containing instances that conform with the interval prediction, i.e., $\calh_{\ell,u} = \cali_{\ell,u} \cap \calh$. 
Then, we formulate an optimization problem to minimize \drcr over all parameterized algorithms under such hard instances.
\begin{subequations}
\label{p:optimization}
\begin{align}
    \min_{\eta, \gamma \ge 1; \bw\in\Omega}\quad& (1-\delta) \eta + \delta \gamma\\
    \label{eq:consist}
    \text{s.t.}\quad& \alg(\bw,I) \le \eta \cdot \opt(I), \forall I \in \calh_{\ell,u},\\
    \label{eq:robust}
    & \alg(\bw,I) \le \gamma \cdot \opt(I), \forall I \in \calh.
\end{align}   
\end{subequations}
Each constraint from either constraint~\eqref{eq:consist} or constraint~\eqref{eq:robust} ensures that the ratio between the expected cost of the online algorithm and the cost of the offline optimum is upper bounded by $\eta$ or $\gamma$, respectively. If restricted only to hard instances $\calh$, the variables $\eta$ and $\gamma$ represent the consistency and robustness of the algorithm $A(\bw)$, and the objective directly optimizes \drcr over all parameterized algorithms. 
Let $\{\eta^*, \gamma^*, \bw^*\}$ denote the optimal solution of the above problem.
Then we propose to choose $A(\bw^*)$ as the online algorithm with UQ prediction $\theta$.

{Since the optimization problem~\eqref{p:optimization} is based on hard instances, its optimal objective provides a lower bound for \drcr over the parameterized algorithms.}

\begin{pro}\label{pro:lb}
No parameterized algorithms $A(\bw), \bw\in\Omega$ can achieve a \drcr smaller than $(1-\delta)\eta^* + \delta \gamma^*$.
\end{pro}
This lower bound can be extended for all online algorithms if the parameterized algorithms can characterize all online algorithms under the hard instances (e.g., see examples in Sections~\ref{sec:sr-pip} and~\ref{sec:os-pip}).
Note that the ancillary problem often involves an infinite number of variables and constraints, which correspond to the high dimension of parameter $\bw$ and the cardinality of hard instance set $\calh$. This necessitates efficient methods for obtaining (approximately) optimal solutions to the problem~\eqref{p:optimization}.
Furthermore, although the optimization can give a lower bound for the target performance, it is essential to additionally establish an upper bound on \drcr of the algorithm $A(\bw^*)$ that is devised based on the solution of the optimization. Developing an online algorithm with matching upper and lower bounds requires carefully constructing the hard instances, crafting the parameterized algorithms, and (approximately) solving the ancillary optimization problem simultaneously.
In Sections~\ref{sec:sr-pip} and~\ref{sec:os-pip}, we showcase how to use this approach to design online algorithms that can make the best use of a given UQ prediction to minimize $\drcr$ in two classic online algorithms problems, the ski rental problem and the online search problem.

}

\section{Ski Rental Problem with UQ Prediction} \label{sec:sr-pip}

\paragraph{Problem statement} 
A player aims to ski for an unknown time horizon $N\in\bbz^+$. Each day she needs to decide whether to rent skis, which cost $\$ 1$ for this day or buy the skis at the cost of $\$ B\in\bbz^+$ and ski for free from then on. 
The goal is to minimize the cost of buying and renting skis.

The difficulty of the problem lies in the uncertain time horizon $N$. If $N$ is known in advance, then the optimal decision is to buy in the beginning if $N \ge B$ and keep renting otherwise. When $N$ is completely unknown, a deterministic online algorithm can achieve a competitive ratio of $2$~\cite{karlin1988competitive}, and this result can be improved to $\nicefrac{e}{e-1}$ by randomization~\cite{karlin1990competitive}. Both results have been proven to be optimal in the worst case.
In previous work on the learning-augmented setting of ski rental, the algorithm is assumed to additionally have access to a deterministic point prediction $P$ over the time horizon $N$ but has no information on the quality of this prediction. There exist both deterministic and randomized algorithms that can attain the Pareto-optimal trade-off between consistency and robustness~\cite{purohit2018improving,wei2020optimal,bamas2020primaldual}. 

We study online algorithms for ski rental with UQ predictions. In particular, UQ about $N$ is given in the form of a probabilistic interval prediction $\theta = \{\ell,u;\delta\}$, i.e., the time horizon $N$ is predicted to be within $\mathbb{Z}^+_{\ell,u} := \{\ell, \ell+1, \dots, u\}$ with at least probability $1-\delta$. 
Instead of making a rent-or-buy decision each day, online decision-making for ski rental can be described as an online (randomized) algorithm with a (random) variable $Y\in\bbz^+$ that keeps renting skis until day $Y-1$ (if the time horizon has not ended) and buys on day $Y$.
We aim to leverage UQ prediction to design the determination of $Y$ so that \drcr can be minimized.
In Section~\ref{sec:sr-ppp}, we first introduce a deterministic algorithm as a warm-up problem to provide insights on algorithms with probabilistic predictions, and then in Section~\ref{sec:ski-rental-opt}, we further propose an optimal randomized algorithm augmented with probabilistic interval predictions using the optimization-based approach.

\subsection{Warm-up: A Deterministic Algorithm}
\label{sec:sr-ppp}
We first focus on a deterministic algorithm for ski rental with a probabilistic point prediction ${\ppp}(P; \delta)$, which forecasts the skiing horizon is $P$ with probability at least $1-\delta$. To simplify the presentation, we show the results based on a continuous version of the ski rental, where the number of skiing days increases continuously, and $N, B, Y \in \mathbb{R}^+$.

\paragraph{A simple meta-algorithm} 
Based on the definition in Equation~\eqref{eq:drcr-trans}, the \drcr of online algorithms is a linear combination of consistency and robustness from an algorithm with untrusted predictions. Therefore, we can devise a simple meta-algorithm by leveraging existing consistent and robust algorithms.
Let $\texttt{LA}_P(\lambda)$ denote the algorithms with untrusted prediction $P$ designed in~\cite{purohit2018improving} for a hyper-parameter ${\lambda\in(0,1]}$.  In particular, $\texttt{LA}_P(\lambda)$ determines the day of purchase $Y = B/\lambda$ if $P < B$ and $Y = B\lambda$ otherwise. $\texttt{LA}_P(\lambda)$ has been proved $(1+\lambda)$-consistent and $(1+1/\lambda)$-robust. 
A simple meta-algorithm then can take $\texttt{LA}_P(\lambda)$ as input and select the online algorithm with parameter $\lambda$
to optimize \drcr. Specifically, it determines $\lambda_{\delta} = \argmin_{\lambda \in (0,1]} (1-\delta)(1+\lambda) + \delta (1 + 1/\lambda) = \min\{\sqrt{\delta/(1-\delta)},1\}$, and the meta-algorithm is given as $\texttt{LA}_P(\lambda_\delta)$. Further, the \drcr of $\texttt{LA}_P(\lambda_\delta)$ is derived as 
\begin{align}
    \label{eq:cr-sr1}
    \crsrppp(\delta) =
    \begin{cases}
        1 + 2\sqrt{\delta(1-\delta)} & \delta \in [0, \frac{1}{2}]\\
        2 & \delta \in (\frac{1}{2}, 1]
    \end{cases}.
\end{align}
The meta-algorithm can improve \drcr beyond the worst-case competitive ratio of $2$ when the prediction quality is high ($\delta \in [0,\nicefrac{1}{2}]$), with $\crsrppp(\delta)$ rapidly converging to $1$ as $\delta$ approaches $0$. 
Nonetheless, the prediction becomes ineffective as its quality deteriorates beyond $\delta > \nicefrac{1}{2}$, reducing the meta-algorithm to the worst-case performance. 
However, a fundamental question remains: \textit{Can we extract the benefit from low-quality predictions?}
Furthermore, the \drcr of the meta-algorithm is independent of the prediction $P$ as the algorithm $\texttt{LA}_P(\lambda)$ treats the prediction $P$ as untrusted and does not leverage its quality $\delta$ in its design. Instead, the prediction quality is only used for the hyper-parameter selection.
These limitations of the meta-algorithm motivate us to design a new algorithm capable of harnessing the probabilistic predictions more effectively.

\begin{algorithm}[t]
\caption{\dsrppp: Deterministic algorithm for ski rental}
\label{alg:ski-ppp1}
\begin{algorithmic}[1]
\State \textbf{input:} prediction $\ppp(P;\delta)$, buying cost $B$;
\State \textbf{if} $P < B$ \textbf{then} determine $Y = B$; 
\State \textbf{else if} {$P\in(\frac{\sqrt{5}+1}{2}B, +\infty)$}  \textbf{then} determine $Y = B \cdot \min\{\sqrt{\delta/(1-\delta)}, 1\}$; 
\State \textbf{else if} {$ P \in [B, \frac{\sqrt{5}+1}{2}B]$} \textbf{then}
\State \quad\textbf{if} $\crsrppp(\delta) \le \delta  + \frac{P}{B}$ \textbf{then} determine $Y = B \cdot \min\{\sqrt{\delta/(1-\delta)}, 1\}$;
\State \quad\textbf{else} determine $Y = P$;
\State buy skis on day $Y$.
\end{algorithmic}
\end{algorithm}

\paragraph{An optimal deterministic algorithm} In Algorithm~\ref{alg:ski-ppp1}, we introduce a new deterministic algorithm, referred to as \dsrppp.
This algorithm operates within distinct prediction regions: (i) in the \textit{pro-rent} region, defined as $P\in(0,B)$, the algorithm purchases on day $B$ regardless of the specific prediction and prediction quality; 
(ii) in the \textit{pro-buy} region, defined as $P\in(\frac{\sqrt{5}+1}{2}B, +\infty)$, the algorithm makes an early purchase within the initial $B$ days, with the specific day determined by the design of \dsrppp; (iii) 
in the \textit{rent-or-buy} region, denoted by $P \in [B, \frac{\sqrt{5}+1}{2}B]$, this algorithm can opt to buy on the predicted day $P$ or make a purchase within the first $B$ days. The decision, in this case, is influenced by both the prediction and its quality, creating a nuanced trade-off between buy and rent. We show that {\dsrppp} can achieve the optimal \drcr among all deterministic algorithms.

\begin{thm}\label{thm:sr-ppp-d}
    Given a $\emph{\ppp}(P; \delta)$, \dsrppp is the optimal deterministic algorithm for ski rental and achieves the \drcr 
    \begin{align*}
    \emph{\drcr(\dsrppp)} = 
    \begin{cases}
        1 + \delta &  P\in(0,B)\\
        \min\left\{\crsrppp(\delta), \delta  + \frac{P}{B} \right\} & P \in [B, \frac{\sqrt{5}+1}{2}B]\\
        \crsrppp(\delta) & P\in(\frac{\sqrt{5}+1}{2}B, +\infty)
    \end{cases}.
    \end{align*}
    Further, \dsrppp achieves the optimal \drcr. 
\end{thm}
The crux of \dsrppp's design lies in identifying the dominant decision when the prediction falls within distinct prediction regions. 
Compared to the meta-algorithm, \dsrppp achieves an improved \drcr over the meta-algorithm for any given prediction. 
The performance gain can be attributed to explicit utilization of both the prediction and its associated quality in the decision-making process. 
In particular, even in scenarios where the prediction quality is low $\delta > \nicefrac{1}{2}$, \dsrppp still manages to enhance the \drcr, especially when the prediction $P < \frac{\sqrt{5}+1}{2}B$. In such cases, any probabilistic information from the prediction can mitigate the worst-case scenarios.
% See Appendix~\ref{app:sr-ppp-d} for a detailed proof. 
{Although we can extend the ideas of \dsrppp to incorporate a probabilistic interval prediction (see Appendix~\ref{app:dsr-pip} for more details), it becomes increasingly complicated to identify the dominant decisions. In the following section, we show that we can design algorithms using a more systematic optimization-based approach proposed in Section~\ref{sec:approach}.
}

\subsection{An Optimal Randomized Algorithm}
\label{sec:ski-rental-opt}

We now introduce a more general randomized algorithm with probabilistic interval prediction $\pip(\ell,u;\delta)$. We can consider a parameterized algorithm $\rsrpip(\by)$ (described in Algorithm~\ref{alg:sr1}) with the purchasing probability $\by := \{y(t)\}_{t\in \mathbb{Z}^+}$ as the parameter. 
Specifically, $y(t)$ denotes the probability of purchasing on day $t$. Then any online randomized algorithm for ski rental problems can be captured by $\by := \{y(t)\}_{t\in \mathbb{Z}^+}$, where $\by$ is a distribution of the buying day with support $\bbz^+$ and the feasible set of $\by$ is given by $\mathcal{Y}:=\{\by: \sum_{t\in\bbz^+} y(t) = 1, y(t) \ge 0, \forall t\in\bbz^+ \}$.

Let $I_N$ denote an instance of the ski rental problem with time horizon $N$. We consider the hard instance set $\calh:= \cali = \{I_N\}_{N\in \bbz^+}$, which in fact contains all instances of the ski rental problem. The instances conforming with the interval prediction can then be denoted by $\calh_{\ell,u} = \{I_N\}_{N\in \bbz^+_{\ell,u}}$.
Given each instance $I_N$, the expected cost of a randomized algorithm $\rsrpip(\by)$ is $\alg(\by,I_N) = \sum\nolimits_{t=1}^{N} (B + t - 1)y(t)  + N\sum\nolimits_{t= N+1}^{+\infty} y(t)$, and the cost of the offline algorithm is $\opt(I_N) = \min\{N,B\}$.
Given a $\pip(\ell,u;\delta)$, we can formulate an optimization problem~\eqref{p:optimization} to minimize \drcr.
Let $\{\eta^*,\gamma^*,\by^*\}$ and $\CR^*_{\texttt{sr}}$ denote the optimal solution and the optimal objective value of the problem, respectively. 
The optimal randomized algorithm is then given by $\rsrpip(\by^*)$.

\begin{algorithm}[t]
\caption{$\rsrpip(\by)$: Randomized algorithm for ski rental}
\label{alg:sr1}
\begin{algorithmic}[1]
\State \textbf{input:} purchase distribution $\by\in\caly$;
 \State draw a buying day $Y$ from the distribution $\by$; 
 \State rent skis up to day $Y-1$ and buy on day $Y$.
\end{algorithmic}
\end{algorithm}

\begin{thm}
\label{thm:sr-rand}
Given a $\emph{\pip}(\ell,u;\delta)$, the \emph{\drcr} of $\rsrpip(\by^*)$ is $\CR^*_{\texttt{sr}}$. Further, $\CR^*_{\texttt{sr}}$ is the optimal \drcr for ski rental.   
\end{thm}

{The optimization-based approach for designing $\rsrpip(\by^*)$ is general in the sense that it can be tuned to design others algorithms for related problems. For example, one can derive a deterministic algorithm with $\pip(\ell,u;\delta)$ by replacing the feasible set with $\hat{\mathcal{Y}} :=\{\by: \sum_{t\in\bbz^+} y(t) = 1, y(t) \in\{0, 1\}, \forall t\in\bbz^+\}$ that restricts the decisions to be deterministic. This systemic design stands in contrast to the ad-hoc development of the deterministic algorithm discussed in the previous section.
Given that $\hat{\mathcal{Y}}\subseteq\caly$, the ancillary problem for the randomized algorithm is a relaxation of that of the deterministic algorithm. Thus, $\rsrpip(\by^*)$ outperforms the optimal deterministic algorithms.  
}

{
Noting that the optimization problem for ski rental is a linear program with an infinite number of variables and constraints, to solve $\by^*$, we show that the problem can be reduced to an equivalent problem with a finite number of variables and constraints. Therefore, $\by^*$ can be solved optimally and efficiently by standard linear programs.
\begin{lem}\label{lem:sr-reduction}
The problem~\eqref{p:optimization} for ski rental can be reduced to an optimization with $O(B)$ variables and $O(B)$ constraints.
\end{lem}
% \bo{A short discussion needed here.}
}

% \vspace{-0.4cm}
\section{Online Search Problem with UQ Prediction}
\label{sec:os-pip}

% \bo{change notations $U\to M$ and $L\to m$, since they conflict with the upper bound function and Lipschitz constant in seection 6.}

% The online search problem, introduced in the seminal work~\cite{el2001optimal}, is another classic online problem under competitive analysis.
% Designing algorithms for online search by leveraging probabilistic predictions requires a more intricate application of the optimization-based algorithmic idea compared to the ski rental problem. 
% We start by considering a special case in Section~\ref{sec:os-ppp} as a warm-up to develop an understanding of the problem, then presenting an optimization-based algorithm for the general setting in Section~\ref{sec:os-pip-opt}.

\paragraph{Problem statement}
A player seeks to sell one unit of a resource over a sequence of prices $\{v_n\}_{n\in[N]}$ that arrive online. In response to each price $v_n$, the player must immediately decide an amount $x_n$ of its remaining resource to sell (resulting in the player earning $v_n x_n$), without the knowledge of future prices or the sequence length $N$. 
If any resource remains unsold at the last step $N$, it is compulsorily sold at the final price $v_N$. The player's goal is to maximize its total profit $\sum_{n\in[N]} v_n x_n$.
% Different assumptions can be made on the feasible set of decisions $\calx \ni x_t$ at each time; we focus on the classic one-way trading~\cite{el2001optimal} problem, where $\calx = [0, 1]$ and the player may sell the item fractionally.
% if $\calx = \{0, 1\}$, then the player must sell the entire resource in a single timestep, and the problem is known as $1$-max search~\cite{el1998competitive}. On the other hand, if $\calx = [0, 1]$, then the player may sell the item fractionally, and the problem is known as one-way trading~\cite{el2001optimal}. We consider both of these settings in our work.
Following the standard assumption~\cite{el2001optimal,lorenz2009optimal}, prices are chosen (possibly adversarially) from a bounded interval, i.e., $v_n \in [m,M]$ for all $n \in [N]$, where $m > 0$~.

% Online search is related to several other online problems that relax the adversarial assumption on prices $v_n$ and instead impose statistical assumptions on the price sequence.
% For instance, the secretary problem~\cite{antoniadis2020secretary} assumes that prices arrive in a uniformly random order, and the Prophet inequality problem~\cite{diakonikolas2021learning} assumes that prices are drawn from a sequence of known distributions. These statistical assumptions can significantly impact the design and analysis of the algorithms. However, they fall beyond the scope of this paper, and we defer exploring them for future research.   

In prior work, there exist several optimal deterministic algorithms (e.g., threat-based algorithm~\cite{el2001optimal}, threshold-based algorithm~\cite{sun2020competitive}) that can achieve the optimal worst-case competitive ratio $\alpha^*=O(\ln(M/m))$. 
Since it is known that randomization does not improve the performance of algorithms for one-way trading problems~\cite{el2001optimal,im2021online}. We focus on deterministic algorithms in this section. 

In online search, if the actual maximum price is known in advance, the offline optimal algorithm simply waits until the maximum price to sell the whole resource.
Previous work on online search with machine-learned advice has considered point predictions of the maximum price \cite{sun2021pareto}. Following this prediction paradigm, in our setting, we consider a probabilistic interval prediction $\pip(\ell,u;\delta)$ of the maximal price, which represents a prediction that the maximum price $V$ lies within the interval $[\ell, u]$ with probability at least $1-\delta$. In the following, we design algorithms to minimize \drcr given predictions of the form $\pip(\ell,u;\delta)$.

% \vspace{-0.2cm}
\subsection{An Optimal-Protection-Function Based Algorithm}
\label{sec:os-pip-opt}

We first introduce a class of ``protection function''-based algorithms (\pfa) in Algorithm~\ref{alg:os2}.
% We first introduce a protection-function-based algorithm (\pfa). 
The $\pfa$ is parameterized by a protection function $G(v):[m,M] \to [0,1]$ that defines the maximum selling amount upon receiving a price $v \in[m,M]$. Then a \pfa only sells the resource if the current price $v_n$ is the maximum one among all previous prices, and the selling amount is $G(v_n) - G(\hat{v})$, where $\hat{v}$ is the previous maximum price.  
$\pfa$ can optimally solve the one-way trading problem when the protection function is given by $G(v) = 0, v\in [m, \alpha^* m)$ and $G(v) = \frac{1}{\alpha^*}\ln\frac{v - m}{\alpha^* m - m}, v\in[\alpha^* m, M]$, where $\alpha^*=O(\ln(M/m))$ is the optimal worst-case competitive ratio. Let $\pfa(G)$ denote the algorithm with protection function $G$.
In the following, we aim to redesign the protection function $G^*$ for \pfa based on the solution of an optimization problem for a given \pip, and show $\pfa(G^*)$ can attain the optimal \drcr.  

\paragraph{Optimization problem based on hard instances.}
We consider hard instances $\calh:= \{I_V\}_{V\in[m,M]}$, where $I_V$ includes a sequence of prices that continuously increase from $m$ to $V$ and then drop to the lowest price $m$ in the end.
Under any instance from $\{I_V\}_{V\in(v,M]}$, $\pfa(G)$ sells $G(v+dv) - G(v)$ amount of resource at price $v$ when the running maximum price increases from $v$ to $v+dv$ for some small $dv$. 
For notational convenience, we define a new parameter $q(v): = [G(v+dv) - G(v)]/dv, \forall v\in[m,M]$. The protection function $G$ can be uniquely determined by 
$\bq: = \{q(v)\}_{v\in[m,M]}$ and the feasible set of $\bq$ is $\calq = \{\bq: q(v) \ge 0, \forall v\in[m,M], \int_{m}^M q(v)dv \le 1\}$. Since the online decision is irrevocable and all instances in $\{I_V\}_{V\in(v, M]}$ have the same prefix (i.e., the price sequence continuously increasing from $m$ to $v$), $q(v)$ is the same for all $\{I_V\}_{V\in (v,M]}$. Moreover, note that any online algorithm corresponds to a solution $\bq :=\{q(v)\}_{v\in[m,M]}$ under the hard instances, and thus we can use $\bq$ to model all online algorithms. Under an instance $I_V$, the profit of an online algorithm modeled by $\bq$ is $\alg(\bq,I_V) = \int_{m}^V v \cdot q(v) dv + (1 - \int_m^V q(v)dv)m$, where the first term is the profit of selling the item over prices from $m$ to $V$ and the second term is the profit from compulsory selling at the last price.
The offline algorithm sells the entire item at the maximum price and thus $\opt(I_V) = V$. 
Given a $\pip(\ell,u;\delta)$, we can formulate an optimization problem~\eqref{p:optimization} to minimize the \drcr under hard instances.
Let $\{\eta^*,\gamma^*, \bq^*\}$ and $\CR^*_{\texttt{os}}$ denote the optimal solution and the optimal objective value. Based on $\bq^*$, we can build a protection function $G^*(v) = \int_m^v q^*(s)ds, \forall v\in[m,M]$ and propose $\pfa(G^*)$ as the algorithm for online search.
  
\begin{algorithm}[t]
\caption{$\pfa(G)$: Protection-function-based algorithm}
\label{alg:os2}
\begin{algorithmic}[1]
\State \textbf{input:} protection function $G$;
\State initiate running maximum price $\hat{v} = m$;
\For{$n = 1,\dots,N-1$}
\State sell $x_n = \left[G^*(v_n) - G^*(\hat{v})\right]^+$;
\State update $\hat{v} = \max\{v_n, \hat{v}\}$;
\EndFor
\State $x_{N} = 1 - G^*(\hat{v})$.
\end{algorithmic}
\end{algorithm}

{
\begin{thm}
\label{thm:os-general}
Given a $\emph{\pip}(\ell,u;\delta)$, the \drcr of $\pfa(G^*)$ is $\CR^*_{\texttt{os}}$, and $\CR^*_{\texttt{os}}$ is optimal for online search. 
\end{thm}
\paragraph{Proof of Theorem~\ref{thm:os-general}.}
Note that $\pfa(G^*)$ only sells the resource when the current price is the running maximum one or when it is the last price. Thus, for any instance $I = \{v_n\}_{n\in[N]}$, we can instead focus on a new instance $I' = \{v'_n\}_{n\in[N'+1]}$, where $\{v'_n\}_{n\in[N']}$ is the $N'$ strictly increasing prices of $I$ and $v'_{N'+1} = v_N$. Thus, we can lower bound the profit of $\pfa(G^*)$ by
\begin{subequations}
\begin{align}
&\alg(\bq^*,I) = \alg(\bq^*,I') \\
& = \sum_{n\in[N']} v'_n \int_{v'_{n-1}}^{v'_{n}} q^*(v)dv + [1-G^*(v'_{N'})]v'_{N'+1}\\
& \ge \sum_{n\in[N']} \int_{v'_{n-1}}^{v'_{n}} v q^*(v)dv + [1-G^*(v'_{N'})]m \\
&\ge \int_{0}^{v'_{N'}} v q^*(v)dv + [1-G^*(v'_{N'})]m.
\end{align}    
\end{subequations}
In addition, we have $\opt(I) = \opt(I') = v'_{N'}$. Since $\bq^*$ is the optimal solution of the optimization problem~\eqref{p:optimization}, we have 
$\alg(\bq^*,I) \ge \frac{\opt(I)}{\eta^*}, \forall v'_{N'} \in [\ell, u]$
 and $\alg(\bq^*,I) \ge \frac{\opt(I)}{\gamma^*}, \forall v'_{N'} \in [m,\ell) \cup (u, M]$. And thus the \drcr of $\pfa(G^*)$ is $(1-\delta) \eta^* + \delta \gamma^* = \CR^*_{\texttt{os}}$. 

Since the parameterized $\pfa(G)$ can capture the performance of all online algorithms under hard instances $\calh$, based on Proposition~\ref{pro:lb}, it is  straightforward to show no online algorithms can achieve a \drcr smaller than $\CR^*_{\texttt{os}}$. 
{
The optimization~\eqref{p:optimization} is a problem with infinite number of variables and constraints. 
To obtain the solution, we propose a discrete approximation to solve it.
Further, if we let $\hat{G}^*$ denote the protection function built based on the solution of the approximation problem, the following lemma shows that $\pfa(\hat{G}^*)$ can achieve a \drcr close to $\pfa(G^*)$. 
\begin{lem}
\label{lem:os-discrete}
For a given parameter $\epsilon > 0$, there exists a discrete approximation problem with $O(\frac{\ln(M/m)}{\ln(1+\epsilon)})$ variables and constraints for the problem~\eqref{p:optimization} of online search. Further, $\pfa(\hat{G}^*)$ can achieve $\drcr \le \CR^*_{\texttt{os}} + \epsilon M/m$.  
\end{lem}
}

}

\section{Learning Algorithms with UQ Prediction}
\label{sec:learning-UQ}

In previous sections, we focused on a \textit{single instance} of an online problem with UQ prediction $\theta$, and designed algorithms to optimize the \drcr, i.e., the expected cost ratio under the worst-case distribution in the ambiguity set built by the UQ prediction. However, in practice, the conditional distribution $\xi_\theta$ is often not the worst-case one. 
Furthermore, the \pip may be imprecise due to non-exchangeability of the data or distribution shift, and alternative notions of UQ may be employed, e.g., see~\cite{Abdar_2021}. For instance, one may approximate the predictive distributions, e.g., through Monte-Carlo methods, but these may be imprecise.
In these cases, it may not be tractable to formulate proper ambiguity sets.
This motivates us to consider the \textit{multi-instance} setting, using online learning to learn the intrinsic correlation between UQ predictions and instance costs as well as to go beyond the \drcr guarantees.

\paragraph{Online learning formulation} 
The idea is to learn the mapping, or policy, from any given UQ prediction to an online algorithm over $T$ rounds.
At the beginning of round $t\in[T]$, we receive a UQ prediction $\theta_t \in \Theta$ about the input instance $I_t$. Then we select an algorithm parameter $\bw_t = \pi_t(\theta_t) \in \Omega$ using a chosen policy $\pi_t \in \Pi:\Theta \mapsto \Omega$, and run the online algorithm $A(\bw_t)$ to execute the instance $I_t$ on the fly.
In the end, we observe the entire instance $I_t$ drawn from the unknown conditional distribution $\xi_{\theta_t}$, and the cost function $f_t := f_t(\bw_t;\theta_t) = \frac{\alg(A(\bw_t),I_t)}{\opt(I_t)}: \Omega \to \mathbb{R}^+$, 
which is the cost ratio of the online algorithm $A(\bw_t)$ and the offline optimal solution under the instance $I_t$.
In our formulation, the goal is to compete against a function $U_t:= U_t(\bw_t;\theta_t)$ that upper bounds the expected cost function $\ex_{\xi_{\theta_t}} [f_t(\bw_t;\theta_t)]$, i.e., $U_t(\bw_t; \theta_{t}) \ge \ex_{\xi_{\theta_t}} [f_t(\bw_t;\theta_t)], \forall \bw_t\in\Omega$. 
This upper bound exhibits certain properties (e.g. Lipschitzness) that will allow one to conduct online learning on it.
We aim to select policies $\{\pi_t\}_{t\in[T]}$, which determine the parameter selection of $\{\bw_t\}_{t\in [T]}$ based on the UQ predictions $\{\theta_t\}_{t\in [T]}$, to minimize the policy regret over $T$ instances $\texttt{PREG}_T$, i.e.,
\begin{align}\label{eq:reg}
 \sum\nolimits_{t\in[T]}[ \ex_{\xi_{\theta_t}} f_t(\pi_t(\theta_t);\theta_t) - U_t(\pi^*(\theta_t);\theta_t)],
\end{align}
where $\pi^* = \argmin_{\pi} \sum_{t\in[T]} U_t(\pi(\theta_t);\theta_t)$.
In general, it is impossible to obtain sublinear policy regret if we do not impose any restrictions on the cost functions with respect to the UQ. This is because the instance of each round can only depend on the newly received UQ but may be unrelated to the past observations. Thus, we consider that there is some \textit{local regularity}  
that encodes the notion that similar UQ predictions should yield similar instance costs.
In this paper, we consider cost functions that are $L$-Lipschitz in $\theta$, i.e., for any $\theta_i,\theta_j \in \Theta$, $\sup_{w \in \Omega}|U_i(w;\theta_i)-U_j(w;\theta_j)| \leq L \cdot \|\theta_i - \theta_j\|.$ 
The goal of the online learning algorithm is to achieve a sublinear regret with respect to \textit{any} cost upper bound function $U_t$ that satisfies the local regularity condition. 
This allows our approach to be \textit{adaptive} to the inherent difficulty of the problem instance: the closer the expected cost $\ex_{\xi_{\theta_t}} f_t$ is to being $L$-Lipschitz, the tighter the cost upper bound $U_t$ will be to the true expected cost, and the more optimally the algorithm will perform. 
Furthermore, the \drcr studied in the previous sections is by definition an upper bound of the expected cost. For certain forms of UQ including \pip predictions, the \drcr is Lipschitz with respect to the UQ.
This means that our approach can at least compete against the optimal \drcr, and outperform them when distributions are not adversarially given.

\paragraph{Algorithms and results}
Using the algorithmic framework in \cite{10.1007/978-3-540-72927-3_36}, one can obtain sublinear policy regret as we defined.  
The main idea for the algorithm is to cover the space of UQ prediction $\Theta$ with an $\epsilon$-net, where an instance of a sublinear regret algorithm (e.g., randomized exponentiated 
gradients algorithm) is assigned to each point in the net. Whenever a UQ prediction falls into one of the $\epsilon$-balls, only the algorithm instance assigned to that ball will be run and updated. Thus, every algorithm instance is only run on similar problem instances. In this way, the algorithm exploits local regularities in the UQ space to achieve improved guarantees. 
Specifically, given that the cost upper bound is $L$-Lipschitz, we show that an $\epsilon$-net based algorithm can guarantee a sublinear policy regret $\Tilde{O}(T^{1-\frac{1}{d+2}})$ for general UQ, where $d$ is the covering dimension of the provided UQs.
To attain this result, we extend the algorithm and regret analysis from \cite{10.1007/978-3-540-72927-3_36} by (i) indicating how the algorithm can exploit local regularities other than just Lipschitz policies for convex functions and (ii) refining the regret analysis for competing against cost upper bounds exhibiting Lipschitzness.
{
As concrete examples, we apply the $\epsilon$-net algorithm to derive policy regret guarantees on \drcr for the ski rental and online search problems with \pip $\theta = \{\ell,u;\delta\}$.
Indeed, under mild conditions the \drcr exhibits Lipschitzness with respect to $\theta$, which we prove using the optimization problems from previous sections.
In particular, for ski-rental we can exploit that $\{\ell,u\}$ are discrete parameters to achieve an improved regret $\Tilde{O}(T^{2/3})$ compared to the general guarantees $\Tilde{O}(T^{4/5})$. For online search, we introduce a discretization on the UQ space to obtain Lipschitzness and achieve regret $\Tilde{O}(T^{4/5})$.
The detailed algorithms and results are in Appendix~\ref{app:beyond-pip}.
}

\begin{figure}
    \centering
\includegraphics[width=0.4\textwidth]{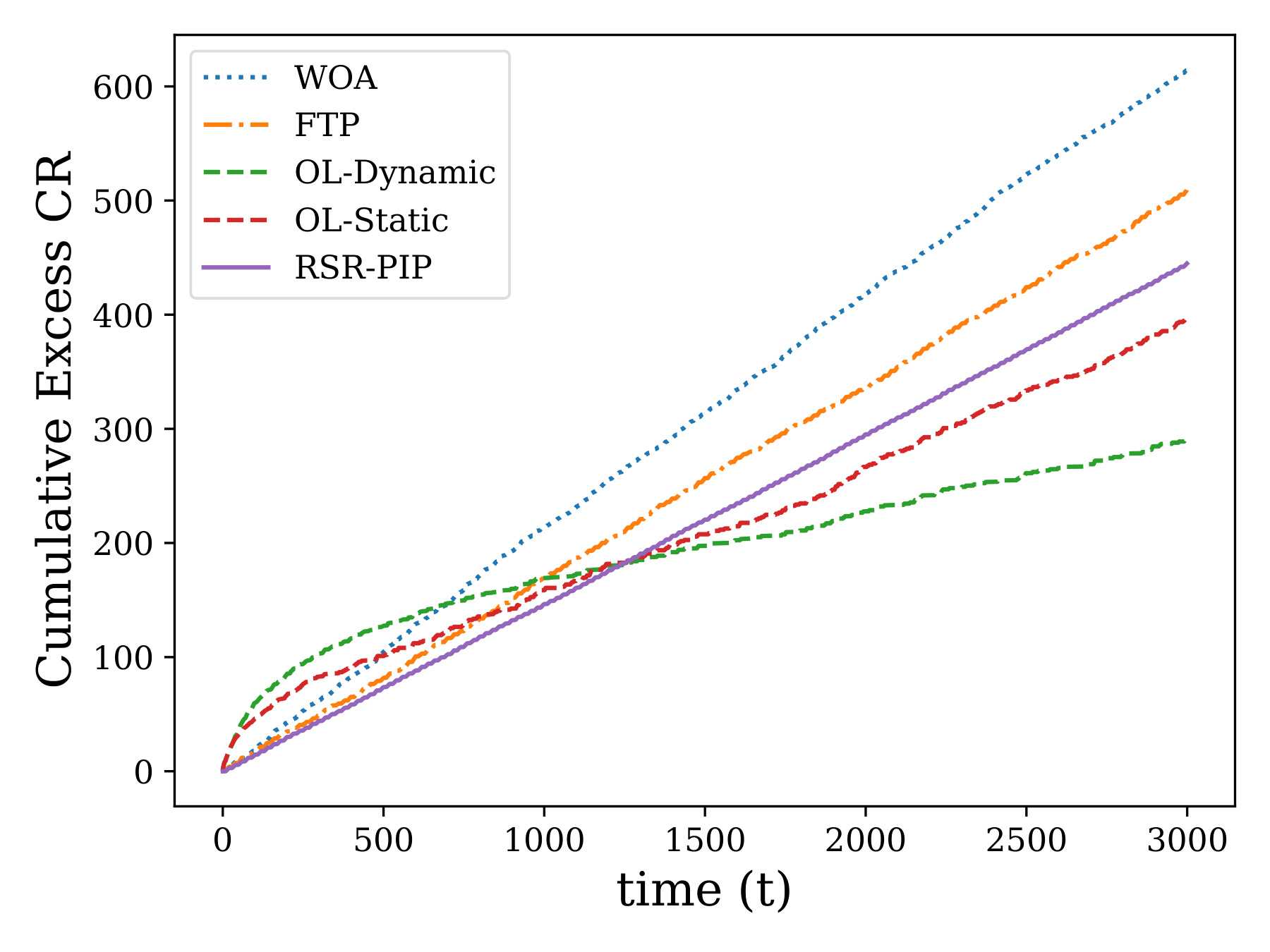}
% \vspace{-0.3cm}
    \caption{ 
    Comparisons of cumulative empirical ratios (minus $1$) of the following algorithms: {\texttt{WOA}}: worst-case optimal randomized algorithm that is $\nicefrac{e}{e-1}$-competitive. {\texttt{FTP}}: follow-the-prediction algorithm that fully trusts the prediction; {\texttt{OL-Dynamic}}: online learning with respect to policy regret by leveraging UQ predictions.
{\texttt{OL-Static}}: online learning with respect to static regret without considering UQ predictions.
\texttt{RSR-PIP}: randomized algorithm with \pip (Algorithm~\ref{alg:os2}) that achieves the optimal \drcr.
    }
    \label{fig:cr}
\end{figure}

\paragraph{Empirical results} 
Figure~\ref{fig:cr} compares the empirical competitive ratios (CRs) of our proposed online algorithms in the setting of a multiple-instance ski rental problem. 
The setup details can be found in Appendix~\ref{app:experiment}.
All our proposed algorithms use UQ to improve the performance compared to those that are worst-case optimized (i.e., \texttt{WOA}) or just use the predictions blindly (i.e., \texttt{FTP}).
Initially, \rsrpip outperforms all other algorithms since \rsrpip is designed to achieve the optimal \drcr, allowing it to perform well before the online learning approaches have had time to learn. 
As the number of instances increases, the cumulative CR of our proposed online learning algorithm \texttt{OL-Dynamic} increases sublinearly, gradually approaching and then outperforming \rsrpip. This is because the distribution used to generate the problem instances is not the worst-case one for the given UQ.  
Thus, \texttt{OL-Dynamic} can better learn to use UQ for non-worst-case distributions, while \rsrpip, designed toward this worst case, performs more conservatively over the long run.
This emphasizes the importance of the online learning approaches in multiple-instance settings in real-world applications, where adversarial distributions rarely occur. 
In addition, the online learning algorithm \texttt{OL-Static}, which is designed for static regret, can also gradually learn to achieve a performance comparable to the optimal \drcr solution but fails to improve much beyond it. This further validates the importance of our policy regret guarantees compared to the classic static regret, which can be obtained without UQ.

\section{Concluding Remarks}
This paper has developed two paradigms for incorporating uncertainty-quantified predictions into the design and analysis of online algorithms.  
For UQ predictions that are descriptive and enable a tractable ambiguity set about the future input to be constructed, we have proposed an optimization-based approach that utilizes the predictions and minimizes a form of distributionally-robust competitive ratio on a per-instance basis. We applied this approach to design optimal online algorithms for two classic online problems with UQ predictions: ski rental and online search problems.
Additionally, we devised an online learning approach that can learn to utilize the predictions across multiple instances and attain sublinear regret under mild Lipschitz conditions. 
We posit that both these paradigms for incorporating uncertainty-quantified advice in online decision-making hold promise for designing algorithms using UQ for other online problems, and can enable better and more reliable use of machine learning in general online decision-making.

\clearpage
\section*{Acknowledgements}
Bo Sun and Raouf Boutaba acknowledge the NSERC Discovery Grant RGPIN-2019-06587.
Jerry Huang is supported by a Caltech Summer Undergraduate Fellowship and the Kiyo and Eiko Tomiyasu SURF Fund. Nicolas Christianson and Adam Wierman acknowledge the support from an NSF Graduate Research Fellowship (DGE-2139433) and NSF Grants CNS-2146814, CPS-2136197, CNS-2106403, and NGSDI-2105648. The work of Mohammad Hajiesmaili is supported by NSF Grants CPS-2136199, CNS-2106299, CNS-2102963, CCF-2325956, and CAREER-2045641.  
We also thank Yiheng Lin and Yisong Yue for insightful discussions. 

\bibliography{reference}
\bibliographystyle{icml2024}

\newpage
\appendix
%%%%%%%%%%%%%%%%%%%%%%%%%%%%%%%%%%%%%%%%%%%%%%%%%%%%%%%%%%%%%%%%%%%%%%%%%%%%%%%
%%%%%%%%%%%%%%%%%%%%%%%%%%%%%%%%%%%%%%%%%%%%%%%%%%%%%%%%%%%%%%%%%%%%%%%%%%%%%%%
% APPENDIX
%%%%%%%%%%%%%%%%%%%%%%%%%%%%%%%%%%%%%%%%%%%%%%%%%%%%%%%%%%%%%%%%%%%%%%%%%%%%%%%
%%%%%%%%%%%%%%%%%%%%%%%%%%%%%%%%%%%%%%%%%%%%%%%%%%%%%%%%%%%%%%%%%%%%%%%%%%%%%%%
\newpage
\appendix
% \onecolumn

% \section{Additional Technical Proofs}
\section{Proof of Proposition~\ref{pro:lb}}

Given a probabilistic interval prediction $\pip(\ell,u;\delta)$, for any parameterized algorithm $A(\bar{\bw}), (\bar{\bw}\in\Omega)$, let $\bar{\eta}$ and $\bar{\gamma}$ denote its consistency and robustness. By definition~\eqref{eq:dfn-rc}, we have
\begin{align*}
    \bar{\eta} = \max_{I \in \cali_{\ell,u}} \frac{\alg(\bar{\bw},I)}{\opt(I)}\ \text{and}\  \bar{\gamma} = \max_{I \in \cali} \frac{\alg(\bar{\bw},I)}{\opt(I)},
\end{align*}
where $\cali$ and $\cali_{\ell,u}$ are the entire instance set and the instance subset that contains all instances confirming with the interval prediction, respectively. Then we have $\calh \subseteq \cali$ and $\calh_{\ell,u} \subseteq \cali_{\ell,u}$ since $\calh$ and $\calh_{\ell,u}$ only contain hard instances. Thus, $\{\bar{\eta},\bar{\gamma}, \bar{\bw}\}$ is a feasible solution of the optimization problem~\eqref{p:optimization}. Consequently, the \drcr of $A(\bar{\bw})$ is $(1-\delta)\bar{\eta} + \delta\bar{\gamma} \ge (1-\delta)\eta^* + \delta\gamma^*$, where $\eta^*$ and $\gamma^*$ are the optimal solution of the problem~\eqref{p:optimization}. Therefore, the \drcr of parameterized algorithms is lower bounded by $(1-\delta)\eta^* + \delta\gamma^*$.

\section{Technical Proofs and Supplementary Results for Ski Rental with UQ Predictions}

\subsection{Proof of Theorem~\ref{thm:sr-ppp-d}}
\label{app:sr-ppp-d}

To design deterministic algorithms for the ski rental problem, we can first derive the distributionally-robust competitive ratio (\drcr) when the buying strategy $Y$ operates in different prediction regions, and then choose the strategy that can minimize the \drcr. 
For notational convenience, we let $\alg$ and $\opt$ denote the cost of the online algorithm and the cost of offline algorithm, and let $\CR$ be the cost ratio of \alg and \opt.  

\noindent{\textbf{Case I}: $P < B$.}
We derive the cost ratios when $Y$ falls in different regions.

\paragraph{Case I(a): when $0 < Y \le P$,}
\begin{itemize}
    \item if $0 < N < Y$, we have $\alg = \opt = N$; and $\CR = 1$;
    \item if $Y \le N < P$, we have $\alg = Y + B$ and $\opt = N$; and thus $\CR = \frac{Y+B}{N}\le \frac{Y+B}{Y}$;
    \item if $N = P$, we have $\alg = Y + B$ and $\opt = P$; and thus $\CR = \frac{Y+B}{P}$;
    \item if $N > P$, we have $\alg = Y + B$ and $\opt = \min\{N,B\}$; and thus $\CR = \frac{Y+B}{\min\{N,B\}}\le \frac{Y+B}{P}$.
\end{itemize}
Thus, we have $\drcr = (1-\delta) \frac{Y+B}{P} + \delta \frac{Y+B}{Y}$, which is minimized by $ Y =
   \begin{cases}
      \sqrt{\frac{\delta}{1-\delta}BP} & \delta \in [0,\frac{P}{P+B}]\\
      P & \delta \in (\frac{P}{P+B},1]
   \end{cases},$
and the corresponding \drcr is $   \begin{cases}
      2\sqrt{\delta(1-\delta)\frac{B}{P}} + (1-\delta) \frac{B}{P} + \delta, & \delta \in [0,\frac{P}{P+B}]\\
      1 + \frac{B}{P} & \delta \in (\frac{P}{P+B},1]
   \end{cases}.$

\paragraph{Case I(b): when $P < Y \le B$,}
\begin{itemize}
    \item if $0 < N < Y$, we have $\alg = \opt = N$; and $\CR = 1$;
    \item if $Y \le N$, we have $\alg = Y + B$ and $\opt = \min\{N,B\}$; and thus $\CR = \frac{Y+B}{\min\{N,B\}}\le \frac{Y+B}{Y}$.
\end{itemize}
Thus, we have $\drcr = 1- \delta + \delta \frac{Y+B}{Y}$, which is minimized when $Y = B$ and $\drcr = 1 + \delta$.

\paragraph{Case I(c): when $Y > B$,} 
\begin{itemize}
    \item if $0 < N \le B$, we have $\alg = \opt = N$; and $\CR = 1$;
    \item if $B < N < Y$, we have $\alg = N$ and $\opt = B$; and thus $\CR = \frac{N}{B}\le \frac{Y}{B}$;
    \item if $ N \ge Y$, we have $\alg = Y+B$ and $\opt = B$; and thus $\CR = \frac{Y+B}{B}$.
\end{itemize}
Thus, we have $\drcr = 1- \delta + \delta \frac{Y+B}{Y}$, which is minimized when $Y = B$ and $\drcr = 1 + \delta$.

By comparing the \drcr of three sub-cases, $Y = B$ minimizes the \drcr when $P < B$, and the minimum \drcr is $1 + \delta$.

\noindent{\textbf{Case II: $ P > B$}.}
Consider the following sub-cases.

\paragraph{Case II(a): when $0 < Y \le B$,}
\begin{itemize}
    \item if $0 < N < Y$, we have $\alg = \opt = N$; and $\CR = 1$;
    \item if $Y \le N \le B$, we have $\alg = Y + B$ and $\opt = N$; and thus $\CR = \frac{Y+B}{N}\le \frac{Y+B}{Y}$;
    \item if $B < N$, we have $\alg = Y + B$ and $\opt = B$; and thus $\CR = \frac{Y+B}{B}$.
\end{itemize}
Thus, we have $\drcr = (1-\delta) \frac{Y + B}{B} + \delta \frac{Y + B}{Y}$, which is minimized when $Y = B \cdot \min\{\sqrt{\delta/(1-\delta)}, 1\}$,
and $\drcr := \crsrppp(\delta) = \begin{cases}
     2\sqrt{\delta(1-\delta)} + 1& \delta \in [0,\frac{1}{2}]\\
     2 & \delta \in (\frac{1}{2},1]
    \end{cases}.$

\paragraph{Case II(b): when $B < Y \le P$,}
\begin{itemize}
    \item if $0 \le N \le B$, we have $\alg = \opt = N$; and $\CR = 1$;
    \item if $B < N < Y$, we have $\alg = N$ and $\opt = B$; and thus $\CR = \frac{N}{B} < \frac{Y}{B}$;
    \item if $Y \le N \le P$, we have $\alg = Y+B$ and $\opt = B$; and thus $\CR = \frac{Y+B}{B} $;
    \item if $P < N$, we have $\alg = Y + B$ and $\opt = B$; and thus $\CR = \frac{Y+B}{B}$.
\end{itemize}
Thus, we have $\drcr = \frac{Y+B}{B}$, which is minimized when $Y \to B$ and $\drcr \to 2$.

\paragraph{Case II(c): when $Y > P$,} 
\begin{itemize}
    \item if $0 < N \le B$, we have $\alg = \opt = N$; and $\CR = 1$;
    \item if $B < N \le P$, we have $\alg = N$ and $\opt = B$; and thus $\CR = \frac{N}{B} \le \frac{P}{B}$;
    \item if $P < N < Y$, we have $\alg = N$ and $\opt = B$; and thus $\CR = \frac{N}{B} < \frac{Y}{B}$;
    \item if $N \ge Y$, we have $\alg = Y+B$ and $\opt = B$; and thus $\CR = \frac{Y+B}{B}$.
\end{itemize}
Thus, we have $\drcr = (1-\delta) \frac{P}{B} + \delta \frac{Y+B}{B}$, which is minimized when $Y \to P$, and $\drcr \to \delta + \frac{P}{B}$.
% \end{proof}

By comparing the \drcr of the above three sub-cases, we have when $P > \frac{\sqrt{5}+1}{2}B$, $\crsrppp(\delta) \le \delta + \frac{P}{B}, \forall \delta \in[0,1]$ and thus, the optimal buying day is $Y = B \min\{\sqrt{\delta/(1-\delta)}, 1\}$. When $B \le P \le \frac{\sqrt{5}+1}{2}B$, the optimal buy strategy is $Y = B \min\{\sqrt{\delta/(1-\delta)}, 1\}$ if $\crsrppp(\delta) \le \delta + \frac{P}{B}$ and $Y = P$ otherwise.

\subsection{Deterministic Algorithms with Probabilistic Interval Predictions}
\label{app:dsr-pip}
We can extend the ideas of \dsrppp to incorporate a probabilistic interval prediction $\pip(\ell,u;\delta)$.
The extension adheres to a decision structure similar to that of \dsrppp when dealing with predicted intervals that clearly favor buying decisions (i.e., $B < \ell \le u$) or renting decisions (i.e., $\ell < u \le B $).  
The additional complexity arises when $\ell \le B \le u$. In such cases, the optimal decision greatly depends on the prediction interval and its quality, and this leads to three possible scenarios: (i) making a pro-rent decision by purchasing at the predicted interval upper bound $u$, (ii) opting for a pro-buy decision by purchasing within the first $\ell$ days , or (iii) buying on day $B$ when the prediction is proved to be unhelpful. The full algorithm is presented in Algorithm~\ref{alg:sr2} and its \drcr result is presented in Lemma~\ref{thm:sr-pip-d}. 
The proof of Lemma~\ref{thm:sr-pip-d} follows the same proof idea as Theorem~\ref{thm:sr-ppp-d}.

\begin{algorithm}[h]
\caption{Online deterministic algorithm with \pip for ski rental}
\label{alg:sr2}
\begin{algorithmic}[1]
\State \textbf{input:} prediction $\pip(\ell,u;\delta)$, buying cost $B$;
\If{$\ell \le u < B$}
\State set $Y = B$;
\ElsIf{$B < \ell \le u$}
\If{$\crsrppp(\delta) \le \delta + \frac{u}{B}$}
\State set $Y = B \cdot \min\{\sqrt{{\delta}/{(1-\delta)}},1\}$;
\Else
\State set $Y = u$;
\EndIf
% \State buy on day $x = B\cdot\hat{u}$
\ElsIf{$\ell \le B \le u$}
\If{$\crsrpip(\delta,\ell) \ge 2$ and $\delta + \frac{u}{B} \ge 2$}
\State set $Y = B$;
\ElsIf{$\crsrpip(\delta,\ell) \le \delta + \frac{u}{B}$}
\State set $Y = \ell\cdot \min\{\sqrt{{B \delta}/{(\ell(1-\delta))}},1\}$;
\Else
\State set $Y = u$;
\EndIf
\EndIf
\State buy skis on day $Y$.
\end{algorithmic}
\end{algorithm}

\begin{thm}\label{thm:sr-pip-d}
Given a $\emph{\pip}(\ell,u;\delta)$, Algorithm~\ref{alg:sr2} for ski rental achieves the \drcr
\begin{align}\label{eq:cr-skirental} 
    \begin{cases}
        1 + \delta & \ell \le u < B\\
        \min\left\{\crsrppp(\delta), \delta  + \frac{u}{B}\right\} & B < \ell \le u\\
        \min\left\{\crsrpip(\delta, \ell), \delta  + \frac{u}{B}, 2\right\} & \ell \le B \le u
    \end{cases}, 
\end{align}
where 
$$\crsrpip(\delta, \ell) := \begin{cases}
        \delta + {(1-\delta)B}/{\ell} + 2\sqrt{{\delta(1-\delta)B}/{\ell}} & \delta \in [0, \frac{\ell}{\ell + B})\\
        1 + {B}/{\ell} & \delta \in [\frac{\ell}{\ell + B}, 1]
    \end{cases}.$$
Further, the attained $\drcr$ is optimal for all deterministic algorithms for ski rental with probabilistic interval predictions.
\end{thm}

\subsection{Proof of Theorem~\ref{thm:sr-rand}}

First, the optimization formulation for the ski rental problem with a probabilistic interval prediction $\pip(\ell,u;\delta)$ can be formally stated as follows: 
\begin{subequations}
\label{p:ski-rental}
\begin{align}
    \min_{\eta, \gamma, \by}\quad& (1-\delta) \eta + \delta \gamma\\
    \label{eq:sr-consist}
    \text{s.t.}\quad& \sum\nolimits_{t=1}^{N} (B + t - 1)y(t)  + N\sum\nolimits_{t= N+1}^{\infty} y(t) \le \eta \min\{N,B\}, \forall N \in \bbz^+_{\ell,u},\\
    \label{eq:sr-robust}
    & \sum\nolimits_{t=1}^{N} (B + t - 1)y(t)  + N\sum\nolimits_{t= N+1}^{\infty} y(t) \le \gamma \min\{N,B\}, \forall N \in \bbz^+\setminus \bbz^+_{\ell,u},\\
    % & \sum\nolimits_{t=1}^{\infty} y(t) = 1,\\ 
    & 1 \le \eta \le \gamma,\\ &\by\in\mathcal{Y}.
\end{align}   
\end{subequations}
Each constraint $N$ from either constraint~\eqref{eq:sr-consist} or constraint~\eqref{eq:sr-robust} ensures that the ratio between the expected cost of the online algorithm and the cost of the offline optimum is upper bounded by $\eta$ or $\gamma$, respectively.
The constraint $1\le \eta \le \gamma$ guarantees that $\sum\nolimits_{t=1}^{N} (B + t - 1)y(t)  + N\sum\nolimits_{t= N+1}^{\infty} y(t) \le \gamma \min\{N,B\}, \forall N \in \bbz^+_{\ell,u}$, that corresponds to the constraint~\eqref{eq:robust} in the optimization problem~\eqref{p:optimization}.

Given any instance $I_N$, the expected cost of $\rsrpip(\by^*)$ is $\alg(\by^*, I_N) = \sum\nolimits_{t=1}^{N} (B + t - 1)y^*(t)  + N\sum\nolimits_{t= N+1}^{\infty} y^*(t)$, and the offline optimal cost is $\opt(I_N) = \min\{N, B\}$. 
Thus, based on the definition in Equation~\eqref{eq:drcr-trans}, $\eta^*$ and $\gamma^*$ are the consistency and robustness of the algorithm with untrusted interval prediction $[\ell, u]$, respectively. And the \drcr of $\rsrpip(\by^*)$ is $(1-\delta)\eta^* + \delta\gamma^* = \CR^*_{\texttt{sr}}$.

To show the optimality of the result, we note that the hard instance set $\calh$ used for formulating the problem is in fact the entire instance set $\cali=\{I_N\}_{N\in\bbz^+}$ for the ski rental problem, and the parameterized algorithms $\rsrpip(\by)$ can capture all online algorithms under $\calh$. Thus, based on Proposition~\ref{pro:lb}, no online algorithms can achieve a \drcr smaller than $\CR^*_{\texttt{sr}}$.

\subsection{Proof of Lemma~\ref{lem:sr-reduction}}

The proof is based on the optimization formulation~\eqref{p:ski-rental}. Let $\calc_N$ denote the constraint indexed by $N$ from the constraints~\eqref{eq:sr-consist} and~\eqref{eq:sr-robust}. Then the problem~\eqref{p:ski-rental} can be reduced to an optimization with a finite number of constraints and variables. Particularly, variables $\{y(t)\}_{t\in\bbz^+}$ and constraints $\{\calc_N\}_{N\in\bbz^+}$ (i.e., constraints~\eqref{eq:sr-consist} and~\eqref{eq:sr-robust}) can be reduced as follows:
\begin{itemize}
\item when $\ell \le u < B$, only $B$ variables $\{y(t)\}_{t\in\bbz^+_{1,B}}$ and $B$ constraints $\{\calc_N\}_{N\in\bbz^+_{1,B}}$  are non-redundant;
\item when $B < \ell \le u$ or $\ell \le B \le u$, only $B+1$ variables $\{y(t)\}_{t\in\bbz^+_{1,B} \cup \{u+1\}}$ and $B+1$ constraints $\{\calc_N\}_{N\in\bbz^+_{1,B-1} \cup \{u, u+1\}}$ are non-redundant.
\end{itemize}

We start by showing the structural property of constraints~\eqref{eq:sr-consist} and~\eqref{eq:sr-robust}. 
Let $\rhs(N)$ and $\lhs(\by, N)$ denote the right-hand-side and the left-hand-side of the constraint $\calc_N$, respectively. 
Given a feasible solution $\{\eta, \gamma, \by\}$, if there exists a $k \in\bbz^+$ such that $\rhs(k) \ge \rhs(k+1)$, we can move the probability mass from $y(k+1)$ to $y(k)$ and obtain a new solution $\hat{\by}$, i.e., 
\begin{align*}
 \hat{y}(t) = 
 \begin{cases}
     y(t) + y(t+1) & t = k\\
     0 &  t = k+1\\
     y(t) & \text{otherwise} 
 \end{cases},
\end{align*}
and the solution $\{\eta,\gamma,\hat{\by}\}$ is also feasible to the problem~\eqref{p:ski-rental}.
To show this, we make the following claims.

\paragraph{\textbf{Claim 1.}} $\{\eta,\gamma,\hat{\by}\}$ satisfies the constraints $\{\calc_N\}_{N\in\bbz^+_{1,k-1}}$.

Note that $\lhs(\by, N) = \sum\nolimits_{t=1}^{N} (B + t - 1)y(t)  + N[1 - \sum\nolimits_{t= 1}^{N} y(t)]$. Then we have $\lhs(\hat{\by},N) = \lhs({\by}, N), \forall N \le k - 1$, and thus the first $k-1$ constraints are feasible for $\hat{\by}$.

\paragraph{\textbf{Claim 2.}} $\{\eta,\gamma,\hat{\by}\}$ satisfies the constraints $\{\calc_N\}_{N\in\bbz^+_{k+1,\infty}}$.

When $N \ge  k+1$, we have
\begin{align*}
  \lhs(\hat{\by}, N) - \lhs({\by}, N) = (B + k - 1) [\hat{y}(k) - y(k)] - (B + k) y(k+1) = - y(k+1) \le 0,  
\end{align*}
and thus all constraints after $k+1$ are feasible for $\hat{\by}$.

\paragraph{\textbf{Claim 3.}} $\{\eta,\gamma,\hat{\by}\}$ satisfies the constraint $\calc_k$.

Note that $\lhs(\hat{\by},k)$ is dominated by $\lhs(\hat{\by},k+1)$ since 
\begin{align*}
   \lhs(\hat{\by},k+1) - \lhs(\hat{\by},k) = 1 - \sum\nolimits_{t=1}^k \hat{y}(t) + (B-1) \hat{y}(k+1) > 0.
\end{align*}
Therefore, if $\rhs(k) \ge \rhs(k+1)$ and $\calc_{k+1}$ is satisfied, $\calc_k$ is also feasible for $\hat{\by}$.

Combining above three claims gives the structural property. 
Next, we show the reduction of variables $\{y(t)\}_{t\in\bbz^+}$ and constraints $\{\calc_N\}_{N\in\bbz^+}$ in the problem~\eqref{p:ski-rental} as follows.

\paragraph{Case I: $\ell \le u < B$.} In this case, $\rhs(N)$ remains the same when $N \ge B$. Therefore, we can iteratively move all the probability mass from $\{y(t)\}_{t\in\bbz^+_{B+1, \infty}}$ to $y(B)$ and obtain the new feasible solution 
\begin{align*}
 \hat{y}(t) = 
 \begin{cases}
     y(t) & t \in\bbz^+_{1,B-1}\\
     \sum_{t=B+1}^\infty y(t) &  t = B\\
     0 & t \in\bbz^+_{B+1,\infty}
 \end{cases}.
\end{align*}

Since $\hat{y}(t) = 0, \forall t \in\bbz^+_{B+1,\infty}$, all variables $\{y(t)\}_{t\in\bbz^+_{B+1,\infty}}$ and constraints $\{\calc_N\}_{N\in\bbz^+_{B+1,\infty}}$ are redundant.
Thus, we only need to focus on $B$ variables $\{y(t)\}_{t\in\bbz^+_{1,B}}$ and $B$ constraints $\{\calc_N\}_{N\in\bbz^+_{1,B}}$.

{
\paragraph{Case II: $B < \ell \le u$.} Note that (i) $\rhs(N) = \gamma 
 B$ remains the same when $N\in \bbz^+_{u+1,\infty}$; and (ii) $\rhs(N)$ is non-increasing in $N$ when $N\in \bbz^+_{B,u}$ since we have $\rhs(N) = \gamma 
 B, N \in \bbz^+_{B,\ell - 1}$ and $\rhs(N) = \eta 
 B, N \in \bbz^+_{\ell,u}$. Based on the structural property, we can iteratively move the probability mass from $\{y(t)\}_{t\in\bbz^+_{u+2, \infty}}$ to $y(u+1)$, and from $\{y(t)\}_{t\in\bbz^+_{B+1, u}}$ to $y(B)$. This gives a new feasible solution  
\begin{align}\label{eq:proof-lem1}
 \hat{y}(t) =
 \begin{cases}
     y(t) & t \in\bbz^+_{1,B-1}\\
     \sum_{t=B}^{u} y(t) & t = B\\
     0 & t=B+1,\dots,u\\
     \sum_{t=u+1}^{\infty} y(t) & t = u+1\\
     0 & t \in\bbz^+_{u+2,\infty}
 \end{cases}.
\end{align}
Thus, we can just focus on the $B+1$ variables $\{y(t)\}_{t\in\bbz^+_{1,B} \cup \{u+1\}}$. 

For $N \in \bbz^+_{B,u}$, note that $\lhs(\hat{\by},N)$ is non-decreasing and $\rhs(N)$ is non-increasing in $N$. Thus, the last constraint $\calc_{u}$ is the most difficulty one and we can just focus on $\calc_{u}$. For $N \in \bbz^+_{u+1,\infty}$, we can just focus on the constraint $\calc_{u+1}$ since $\hat{y}(t) = 0, \forall t\in \bbz^+_{u+2,\infty}$. Thus, we only need to consider the constraints $\{\calc_N\}_{N\in\bbz^+_{1,B-1} \cup \{u, u+1\}}$.

\paragraph{Case III: $\ell \le B \le u$.} 
In this case, we have that (i) $\rhs(N) = \gamma 
 B$ remains the same when $N\in \bbz^+_{u+1,\infty}$; and (ii) $\rhs(N) = \eta 
 B$ remains the same when $N\in \bbz^+_{B,u}$. Then, the probability mass $\{y(t)\}_{t\in\bbz^+_{u+2, \infty}}$ can be moved to $y(u+1)$, and the probability mass $\{y(t)\}_{t\in\bbz^+_{B+1, u}}$ can be moved to $y(B)$. Thus, a new feasible solution $\hat{\by}$ is also given in the same form as Equation~\eqref{eq:proof-lem1} and we can focus on the $B+1$ variables $\{y(t)\}_{t\in\bbz^+_{1,B} \cup \{u+1\}}$. 

For $N \in \bbz^+_{B,u}$, $\lhs(\hat{\by},N)$ is non-decreasing and $\rhs(N)$ is constant in $N$. We can just focus on the last constraint $\calc_{u}$. For $N \in \bbz^+_{u+1,\infty}$, we can just focus on the constraint $\calc_{u+1}$ since $\hat{y}(t) = 0, \forall t\in \bbz^+_{u+2,\infty}$. Thus, we only need to consider the constraints $\{\calc_N\}_{N\in\bbz^+_{1,B-1} \cup \{u, u+1\}}$.

Combining \textit{Case I} - \textit{Case III}, there are no more than $B+1$ non-redundant variables in $\{y(t)\}_{t\in\bbz^+}$, and $B+1$ constraints in $\{\calc_N\}_{N\in\bbz^+}$. This completes the proof.

}

%%%%%%%%%%%%%%%%%%%%%%%%%%%%%%%%%%%%%%%%%%%%%%%%%%%%%%%%%%%%%%%%%%%%%%%%%%%%%%%%%%%%%%%%%%%%%%%%%%%%%%%%%%%%%%%%%%%%%%%%%%5
\section{Technical Proofs and Supplementary Results for Online Search with UQ Predictions}

\subsection{Discrete Approximation for the Optimization Problem}
\label{app:discrete-approx}

We start by providing a formal formulation for the online search with a probabilistic interval prediction $\pip(\ell,u;\delta)$ under the hard instances $\calh :=\{I_V\}_{V\in[m,M]}$.
\begin{subequations}
\label{p:online-search}
\begin{align}
    \min_{\eta,\gamma, \bq} \quad& (1-\delta) \eta + \delta \gamma\\
    {\rm s.t.}\quad& V \le \eta \left[\int_{m}^V v \cdot q(v) dv + (1 - \int_m^V q(v)dv)m \right], V\in [\ell,u], \label{eq:constr1}\\
    & V \le \gamma \left[\int_{m}^V v \cdot q(v) dv + (1 - \int_m^V q(v)dv)m \right], V\in [m,\ell) \cup (u,M], \label{eq:constr2}\\
    & 1 \le \eta \le \gamma, \\
    &\bq \in \calq.
\end{align}    
\end{subequations}
Each constraint indexed by $V$ in Equation~\eqref{eq:constr1} (or in Equation~\eqref{eq:constr2}) ensures the ratio between the profit of the offline optimum and the profit of the online algorithm is upper bounded by $\eta$ (or $\gamma$) under the instance $I_V$. Thus, $\eta$ and $\gamma$ represent the consistency and robustness (under the hard instances), and the optimization objective is to minimize the \drcr under the hard instances.
Let $\{\bq^*,\eta^*,\gamma^*\}$ and $\CR^*_{\texttt{os}}$ denote the optimal solution and the optimal objective value of the above problem.

We propose a discrete approximation for the problem~\eqref{p:online-search} as follows. Fix a parameter $\epsilon > 0$. Let $K'$ be the largest integer such that $m (1+\epsilon)^{K'} \le M$, i.e., $K' = \lfloor \frac{\ln(M/m)}{\ln(1+\epsilon)} \rfloor$. 
Consider the following $K = K' + 4$ discrete values that include $K'+1$ values $\{m(1+\epsilon)^k\}_{k=0,\dots,K'}$, and three additional values $\ell$, $u$, $M$. We arrange these $K$ values in a non-decreasing order and let $V_k$ denote the $k$-th value. Particularly, we have $V_1 = m, V_{K} = M$, and we define index $k_\ell$ and $k_u$ such that $V_{k_\ell} = \ell$ and $V_{k_u} = u$.
The key property of the defined discrete values is that ${V_{k}}/{V_{k-1}} \le 1+\epsilon, \forall k = 2,\dots, K$.
We define $K$ variables $\hat{\bq}:=\{\hat{q}_k\}_{k\in[K]}$ and its feasible set $\hat{\calq} :=\{\hat{\bq}: \hat{q}_k \ge 0, \forall k\in[K], \sum_{k\in[K]}\hat{q}_k \le 1\}$.
Then we consider the following discrete version of the problem~\eqref{p:online-search}.
\begin{subequations}
\label{p:online-search-d}
\begin{align}
    \min_{\hat{\eta},\hat{\gamma}, \hat{\bq} }\quad& (1-\delta) \hat{\eta} + \delta \hat{\gamma}\\
    \label{eq:os-consist-d}
    {\rm s.t.}\quad& V_k \le \hat{\eta} \left[\sum\nolimits_{i=1}^{k} V_{i} \hat{q}_i + (1 - \sum\nolimits_{i=1}^{k} \hat{q}_i) m \right], k = k_\ell, \dots, k_u, \\
    \label{eq:os-robust-d}
    & V_k \le \hat{\gamma} \left[\sum\nolimits_{i=1}^{k} V_{i} \hat{q}_i + (1 - \sum\nolimits_{i=1}^{k} \hat{q}_i) m \right], k = 1, \dots, k_\ell - 1, k_u + 1, \dots, K,\\
    & 1 \le \hat{\eta} \le \hat{\gamma},\\
    &\hat{\bq} \in \hat{\calq},
\end{align}    
\end{subequations}
which only contains $K$ variables in $\hat{\bq}$ and $K$ constraints in Equations~\eqref{eq:os-consist-d} and~\eqref{eq:os-robust-d}.

\subsection{Proof of Lemma~\ref{lem:os-discrete}}
\label{app:os-discrete-approx}

Based on the discrete approximation~\eqref{p:online-search-d} proposed in Appendix~\ref{app:discrete-approx}, we can have an approximate problem with $O(\frac{\ln(M/m)}{\ln(1+\epsilon)})$ variables and constraints.
Below we prove the approximation error of $\pfa(\hat{G}^*)$.

First, we show the discrete problem~\eqref{p:online-search-d} is a relaxation of the original problem~\eqref{p:online-search}. The relaxation is by (i) removing all constraints except when $V$ takes the $K$ discrete values $\{V_k\}_{k\in[K]}$; and (ii) further relaxing the remaining $K$ constraints as follows. 
The $K$ remaining constraints of the original problem can be shown as 
\begin{subequations}
\begin{align}
\label{eq:os-consist-c}
&V_k \le \hat{\eta} \left[\int_{m}^{V_k} v \cdot \hat{q}(v) dv + (1 - \int_m^{V_k} \hat{q}(v)dv)m \right], k = k_\ell, \dots, k_u, \\
\label{eq:os-robust-c}
& V_k \le \hat{\gamma} \left[\int_{m}^{V_k} v \cdot \hat{q}(v) dv + (1 - \int_m^{V_k} \hat{q}(v)dv)m \right], k = 1, \dots, k_\ell - 1, k_u + 1, \dots, K.
\end{align}    
\end{subequations}
Constraints~\eqref{eq:os-consist-d} and constraints~\eqref{eq:os-robust-d} are further relaxed constraints of the constraints~\eqref{eq:os-consist-c} and constraints~\eqref{eq:os-robust-c}, respectively.
In particular, we relax $\int_{m}^{V_k} v \cdot \hat{q}(v) dv$ to $\sum_{i=1}^k V_i \hat{q}_i, \forall k\in[K]$.

Let $\{\hat{\eta}^*, \hat{\gamma}^*, \hat{\bq}^*\}$ denote the optimal solution of the discrete problem~\eqref{p:online-search-d}.
Since the discrete problem is a relaxation of the original problem, we have $(1-\delta) \hat{\eta}^* + \delta \hat{\gamma}^* \le (1-\delta) \eta^* + \delta \gamma^*$.

Based on $\hat{\bq}^*$, we can build a piece-wise constant protection function $\hat{G}^*(v) = \sum_{i=1}^k \hat{q}_i$ if $v \in [V_{k}, V_{k+1}], k\in[K-1]$.
We then aim to analyze the \drcr of $\pfa(\hat{G}^*)$.
Following similar approach to the proof of Theorem~\ref{thm:os-general},  $\pfa(\hat{G}^*)$ can guarantee
\begin{align*}
    \alg(\hat{\bq}^*,I_{v'_{N'}}) &\ge \frac{\opt(I_{v'_{N'}})}{\frac{V_{k}}{V_{k-1}}\hat{\eta}^*} \ge  \frac{\opt(I_{v'_{N'}})}{(1+\epsilon)\hat{\eta}^*}, \forall v'_{N'} \in [\ell, u]\\
    \alg(\hat{\bq}^*,I_{v'_{N'}}) &\ge \frac{\opt(I_{v'_{N'}})}{\frac{V_{k}}{V_{k-1}}\hat{\gamma}^*} \ge \frac{\opt(I_{v'_{N'}})}{(1+\epsilon)\hat{\gamma}^*}, \forall v'_{N'} \in [m,\ell) \cup (u, M].
\end{align*}
Thus, the \drcr of $\pfa(\hat{G}^*)$ is $(1+\epsilon)[(1-\delta) \hat{\eta}^* + \delta \hat{\gamma}^*] \le (1+\epsilon)[(1-\delta)\eta^* + \delta \gamma^*] \le (1-\delta)\eta^* + \delta \gamma^* + \epsilon M/m$. This completes the proof.

%%%%%%%%%%%%%%%%%%%%%%%%%%%%%%%%%%%%%%%%%%%%%%%%%%%%%%%%%%%%%%%%

\section{Technical Proofs and Supplementary Results for Learning Algorithms with UQ Predictions}
\label{app:beyond-pip}

\subsection{Algorithm and Main Results} \label{sec:col}

The key algorithmic framework we use is based on~\cite{10.1007/978-3-540-72927-3_36}, as illustrated in Algorithm~\ref{alg:col}. 
Given that there exists a master algorithm that can achieve $\Tilde{O}(\sqrt{T})$ static regret on the cost sequence $\{f_t\}_{t\in[T]}$ and the cost upper bounds $\{U_t\}_{t\in[T]}$ are Lipschitz in UQ predictions, we show that one can use Algorithm~\ref{alg:col}
to achieve a sublinear policy regret $\texttt{PREG}_T$. We prove this fact in the following theorem.
\begin{thm} \label{thm:lipschitz cost}
For each UQ prediction $\theta_t\in\Theta \subseteq [0,1]^{D}$, suppose cost function $f_t \sim \xi_{\theta_t}$, and that there exists an algorithm $\mathcal{A}$ that achieves $\Tilde{O}(\sqrt{T})$ static regret in expectation for $f_1,...,f_T$. Moreover, assume that the covering dimension of the $\{\theta_t\}_{t\in T}$ is $d \leq D$, i.e., the UQ vectors originate from a $d$-dimensional subspace of $[0,1]^D$. For any cost upper bound $U_t$ of $\mathbb{E}_{\xi_{\theta_t}}f_t$, i.e. $U_t(\bw_t;\theta_t)\geq \mathbb{E}_{\xi_{\theta_t}}f_t(\bw_t;\theta_t)$ for all $\bw_t \in \Omega$, such that $U_t$ is $L$-Lipschitz in $\theta_t \in \Theta$, Algorithm~\ref{alg:col} with $\mathcal{A}$ as its master algorithm achieves the expected policy regret with respect to the optimal policy $\pi^*$ for $U_1,...,U_T$
\begin{align*}
\sum\nolimits_{t\in[T]} [\mathbb{E}f_t(\pi_t(\theta_t);\theta_t)-U_t(\pi^*(\theta_t);\theta_t)] = \Tilde{O}(L^{1-\frac{2}{d+2}}T^{1-\frac{1}{d+2}}),
\end{align*}
    where $\pi^* = \argmin_{\pi} \sum_{t\in[T]} U_t(\pi(\theta_t);\theta_t)$. Here, the expectation is taken over $f_t \sim \xi_{\theta_t}$ and the randomness of the master algorithm.
\end{thm}
By definition, \drcr is a natural cost upper bound for cost functions. However, we remark that in many instances there will likely be a tighter upper bound than the $\drcr$ that is also Lipschitz in $\theta$, and we showed that Algorithm~\ref{alg:col} can automatically compete against any such upper bound. Thus, we expect that in practice, Algorithm~\ref{alg:col} can potentially do better than the bounds stated in this section, and in particular it likely can exploit $\pip$s \textit{more optimally} than optimization-based algorithms. 
We confirm this in our numerical experiments.

\begin{algorithm}[t]
\caption{Online learning algorithm with uncertainty-quantified predictions}
\label{alg:col}
\begin{algorithmic}[1]
\State \textbf{input:} master algorithm $\mathcal{A}$ (e.g., exponentiated 
gradients algorithm); UQ space $\Theta$; parameterized algorithms $A(\bw), \bw\in\Omega$; parameter $\epsilon$;
\State initialize $\epsilon$-net $\mathcal{N}=\emptyset$;
\For{each round $t=1,...,T$}
\State receive UQ $\theta_t \in \Theta$ and let $\Tilde{\theta}_t=\arg\min_{\theta \in \mathcal{N}} \|\theta_t-\theta\|$ be the closest vector in $\mathcal{N}$;
\If{$\|\theta_t-\Tilde{\theta}_t\| > \epsilon$}
\State add $\theta_t$ to $\mathcal{N}$;
\State start a new instance of algorithm $\mathcal{A}_{{\theta}_t}$ corresponding to ${\theta}_t$;
\State update $\Tilde{\theta}_t \leftarrow {\theta}_t$;
\EndIf
\State choose $\bw_t$ as the output of $\mathcal{A}_{\Tilde{\theta}_t}$;
\State run algorithm $A(\bw_t)$ to execute the instance $I_t$, and observe the cost function $f_t(\bw_t;\theta_t)$;
\State update $\mathcal{A}_{\Tilde{\theta}_t}$ based on $f_t(\bw_t;\theta_t)$.
\EndFor
\end{algorithmic}
\end{algorithm}

\subsection{Applications to Ski Rental and Online Search Problems}
\label{sec:learning-app}

We consider how one can use a variant of Algorithm~\ref{alg:col} with master algorithm being the randomized exponentiated (sub)gradient (EG) algorithm~\cite{10.1561/2200000018, mcmahan2015survey} to derive policy regret guarantees on the $\drcr$ for the ski rental and online search problems when UQs are probabilistic interval predictions
$\pip(\theta) = \pip(\ell,u;\delta)$.
This is done by observing that the $\drcr$ upper bounds the expected competitive ratio, and, under mild conditions, \drcr exhibits Lipschitzness with respect to $\theta$. Then, using the ideas in Theorem~\ref{thm:lipschitz cost}, we show that it is sufficient to learn the low-regret policy with respect to $\drcr$ by just using the realized cost ratio function $f_t$ for each instance $t$. 
Also note that the learning algorithm is unaware of $\theta$'s status as UQ, instead treating the vectors $\theta_t$ as generic side information. 
However, by leveraging the structure of the information space for \pip, we can improve the regret guarantees for ski-rental (see Corollary~\ref{cor:dsr-ol}) compared to the general guarantees in Theorem~\ref{thm:lipschitz cost}, and obtain provable Lipschitzness guarantees and, hence, regret guarantees for online search (see Corollary~\ref{cor:os-ol}).

\subsubsection{Ski Rental Problem}
Consider an online sequence of $T$ instances of the ski rental problem, where we assume that $\bar{N} \geq 2$ bounds the number of skiing days and $B > 0$ is the buying cost. We will also assume that at the start of instance $t$, we receive a $\pip(\ell_t,u_t;\delta_t) \in [\bar{N}] \times [\bar{N}] \times [0,1]$.
The Algorithm~\ref{alg:col-app} we use is a slight modification of Algorithm~\ref{alg:col}, where we consider the fact that $\ell_t,u_t$ are discrete. Specifically, we separately consider the spaces $(\ell,u,\cdot)=[0,1]$ indexed by all $\bar{N}^2$ combinations of $(\ell,u)$ (the number of combinations can be further reduced in practice by noting $u \ge \ell$), and we separately cover each space with an $\epsilon$-net $\mathcal{N}_{(\ell,u)}$. Whenever we receive a UQ vector $\theta_t=(\ell_t,u_t;\delta_t)$, we will run the online learning algorithm corresponding to the closest point in $\mathcal{N}_{(\ell_t,u_t)}$.

Let $F_t:=F_t(Y_t;\theta_t) = \frac{\alg(Y_t,I_t)}{\opt(I_t)}: [\bar{N}] \to \mathbb{R}^+$ denote the cost ratio function for ski rental that can be constructed after observing the instance $I_t$, where $F_t(Y_t;\theta_t)$ is the cost ratio of buying on day $Y_t \in [\bar{N}]$. Let $f_t:= f_t(\by_t;\theta_t)$ denote the expected cost ratio function over the randomized decision $Y_t$, i.e., $f_t(\by_t;\theta_t)=\mathbb{E}_{Y_t \sim \by_t} F_t(Y_t;\theta_t)$. We will also denote the $\drcr$ function we derived in Section~\ref{sec:sr-pip} as $U_t:= U_t(\by_t;\theta_t)$, which upper bounds $\ex_{\xi_\theta} f_t$, i.e., $U_t(\by_t;\theta_t) \ge \ex_{\xi_\theta} f_t(\by_t;\theta_t), \forall \by_t \in \mathcal{Y}$, where $\mathcal{Y}$ is the simplex over support $[\bar{N}]$.
By using similar proof ideas to Theorem~\ref{thm:lipschitz cost}, we can show that learning using the cost ratio $f_t$, which we can construct in hindsight after each instance $I_t$, allows us to compete against the $\drcr$ $U_t$.

\begin{cor} \label{cor:dsr-ol}
For the multi-instance ski rental problem, there is a policy $\{\pi_t\}_{t\in[T]}$ that can compete against the optimal policy $\pi^*$ that maps $\Theta$ to the simplex $\mathcal{Y}$ over $[\bar{N}]$ with respect to $\drcr$ $\{U_t\}_{t\in[T]}$, obtaining expected policy regret
\begin{align*}
\sum\nolimits_{t\in[T]} [\mathbb{E} f_t(\pi_t(\theta_t);\theta_t)-U_t(\pi^*(\theta_t);\theta_t)] = \Tilde{O}(\bar{N}(\max\{(\bar{N}+B)/B, B\})T^{2/3}),
\end{align*}
where $\pi^* = \argmin_{\pi} \sum_{t\in[T]} U_t(\pi(\theta_t);\theta_t)$. The expectation is taken over the randomness of the instance distribution and the algorithm. The $\Tilde{O}(\cdot)$ hides factors of $\log \bar{N}$.
\end{cor}

\begin{algorithm}[t]
\caption{Online learning algorithm for multiple-instance ski rental}
\label{alg:col-app}
\begin{algorithmic}[1]
\State \textbf{input:} master algorithm $\mathcal{A}$; space of probabilistic interval prediction $\Theta:= [\bar{N}] \times [\bar{N}] \times [0,1]$; parameterized algorithms $\rsrpip(\by), \by\in\caly$; parameter $\epsilon$;
\State Initialize $\mathcal{N}=\{\mathcal{N}_{(i,j)}=\emptyset\}_{(i,j)\in [\bar{N}] \times [\bar{N}]}$;
\For{each $t=1,...,T$}
\State receive $\theta_t = (\ell_t,u_t;\delta_t) \in [\bar{N}] \times [\bar{N}] \times [0,1]$ and let $\Tilde{\theta}_t=\arg\min_{\theta \in \mathcal{N}_{(\ell_t,u_t)}} \|\theta_t-\theta\|$ be the closest vector in $\mathcal{N}_{(\ell_t,u_t)}$;
\If{$\|\theta_t-\Tilde{\theta}_t\| > \epsilon$}
\State add $\theta_t$ to $\mathcal{N}_{(\ell_t,u_t)}$;
\State start a new instance of algorithm $\mathcal{A}_{{\theta}_t}$ corresponding to ${\theta}_t$;
\State update $\Tilde{\theta}_t \leftarrow {\theta}_t$;
\EndIf
\State choose $\by_t$ as the output of $\mathcal{A}_{\Tilde{\theta}_t}$;
\State run algorithm $\rsrpip(\by_t)$ to execute the instance $I_t$;
\State update $\mathcal{A}_{\Tilde{\theta}_t}$.
\EndFor
\end{algorithmic}
\end{algorithm}

\subsubsection{Online Search Problem}
Consider an online sequence of $T$ instances of the online search problem with prices bounded within $[m,M]$. At the start of instance $I_t$, we receive a $\pip(\ell_t, u_t; \delta_t) \in [m,M] \times [m,M] \times [0,1]$. We use a slight modification of Algorithm~\ref{alg:col-app} to solve this problem, but we further discretize the space of possible $(\ell_t,u_t) \in [m,M]^2$ before applying the algorithm.

Denote $f_t:= f_t(\bq_t;\theta_t) = \frac{\opt(I_t)}{\alg(\bq_t,I_t)}: \calq \to \mathbb{R}^+ $ as the profit ratio function, where $f_t(\bq;\theta_t)$ is the profit ratio for choosing $\bq\in\mathcal{Q}=\{\bq: q(v) \geq 0, v\in[m,M], \int_m^M q(v) dv = 1\}$.
The $\drcr$ is again denoted as $U_t:= U_t(\bq_t;\theta_t)$, which upper bounds $\ex_{\xi_{\theta_t}} f_t(\bq;\theta_t)$. We can design an online learning algorithm that can compete against the optimal $\drcr$.

\begin{cor} \label{cor:os-ol}
For the multi-instance online search problem, there is a policy $\{\pi_t\}_{t\in[T]}$ that can compete against the optimal policy $\pi^*$ mapping $\Theta$ to $\mathcal{Q}$ with respect to $\drcr$ $\{U_t\}_{t\in[T]}$, obtaining expected policy regret
    \begin{equation*}
\sum\nolimits_{t\in[T]} [\mathbb{E}f_t(\pi_t(\theta_t);\theta_t)-U_t(\pi^*(\theta_t);\theta_t)] = \Tilde{O}((M-m+1)((M/m)^{8/5}+1) T^{4/5}),
    \end{equation*}
    where $\pi^* = \argmin_{\pi} \sum_{t\in[T]} U_t(\pi(\theta_t);\theta_t)$. The expectation is taken over the instance distribution. The $\Tilde{O}(\cdot)$ hides factors of $\log(M/m)$, $\log(M-m)$, and $\log T$.
\end{cor}

\subsection{Proof of Theorem~\ref{thm:lipschitz cost}}

For each round $t$, let $\bw_t = \pi_t(\theta_t)$ be the decision of 
Algorithm~\ref{alg:col} and $\bu_t = \pi^*(\theta_t)$ be the offline optimal decision for $U_t$. The policy $\pi_t(\theta_t) = \pi_t(\theta_t|\theta_1,f_1,...,\theta_{t-1},f_{t-1})$ determined by Algorithm~\ref{alg:col} depends on all past observed cost functions $f_1,\dots,f_{t-1}$ and UQ predictions $\theta_1,\dots,\theta_{t-1}$. 
We can reformulate the policy regret defined in~\eqref{eq:reg} as follows:
\begin{align}
\sum\nolimits_{t\in[T]} [\mathbb{E}f_t(\pi_t(\theta_t);\theta_t)-U_t(\pi^*(\theta_t);\theta_t)] = \mathbb{E}\sum\nolimits_{\theta \in \mathcal{N}} \sum\nolimits_{t\in \mathcal{T}_{\theta}}[f_t(\bw_t;\theta_t)-U_t(\bu_t;\theta_t)],    
\end{align}
where $\mathcal{T}_{\theta} = \{t\in [T]: \Tilde{\theta}_t=\theta\}$ denotes set of rounds with UQ corresponding to $\theta \in \mathcal{N}$. Let $T_\theta=|\mathcal{T}_{\theta}|$ be the number of such rounds. We can further bound this quantity as follows:
\begin{subequations}
\begin{align}
\mathbb{E}\sum_{t \in \mathcal{T}_{\theta}}[f_t(\bw_t;\theta_t)-U_t(\bu_t;\theta_t)] &= \mathbb{E}\sum_{t \in \mathcal{T}_{\theta}} [f_t(\bw_t;\theta_t)-f_t(\bu_1;\theta_t)] + \sum_{t \in \mathcal{T}_{\theta}} [\ex f_t(\bu_1;\theta_t)-U_t(\bu_t;\theta_t)]\\
&\leq \Tilde{O}(\sqrt{T_\theta}) + \sum_{t \in \mathcal{T}_{\theta}} [U_t(\bu_1;\theta_t)-U_t(\bu_t;\theta_t)]\\
&\leq \Tilde{O}(\sqrt{T_\theta}) + \sum_{t \in \mathcal{T}_{\theta}} [U_t(\bu_1;\theta_t)-U_1(\bu_1;\theta_1)+U_1(\bu_t;\theta_1)-U_t(\bu_t;\theta_t)]\\
&\leq \Tilde{O}(\sqrt{T_\theta}) + 2T_\theta \cdot \epsilon L ,
\end{align}  
\end{subequations}
where the first inequality follows since the master algorithm $\mathcal{A}$ guarantees the static regret and $U_t(\bu_1;\theta_t)$ upper bounds the expected cost $\ex f_t(\bu_1;\theta_t)$, the second inequality follows since $\bu_1$ is the minimizer of $U_1(\bu;\theta_1)$, and the last one follows from our Lipschitz assumption of the cost upper bounds.

Then, using the Cauchy-Schwarz inequality, 
\begin{equation}
\sum_{\theta \in \mathcal{N}}\sum_{t \in \mathcal{T}_{\theta}} [\ex f_t(\bw_t;\theta_t)-U_t(\bu_t;\theta_t)] \leq \sum_{\theta \in \mathcal{N}} \Tilde{O}(\sqrt{T_\theta})+2T_\theta \epsilon L \leq \Tilde{O}(\sqrt{|\mathcal{N}|T}) + 2T\epsilon L.
\end{equation}
    Since the covering dimension is $d$ and the distance between any two UQ points in the net $\mathcal{N}$ constructed by \Cref{alg:col} is $\epsilon$, one can deduce by volume arguments that the $\epsilon$-net has size $|\mathcal{N}|=O(1/\epsilon^d)$. Finally, choosing $\epsilon=(L^2T)^{-1/(d+2)}$ minimizes this expression, giving the final result.
% \end{proof}

\subsection{Proof of Corollary~\ref{cor:dsr-ol}}
\label{app:dsr-ol-proof}
We use Algorithm~\ref{alg:col-app} with the master algorithm being the randomized exponentiated (sub)gradient (EG) algorithm~\cite{10.1561/2200000018, mcmahan2015survey}, which is an algorithm for learning from experts.
EG runs on the simplex $\mathcal{Y}$ over $[\bar{N}]$ and the function $f_t$, in order to learn a randomized decision $\by\in \mathcal{Y}$, from which the decision $Y$ is sampled from.
Note that $f_t$ is bounded above by $\max\{(\bar{N}+B)/B, B\}$. Thus, by setting the step size to be $\frac{\sqrt{\log \bar{N}}}{\max\{(\bar{N}+B)/B, B\}\sqrt{2T}}$ and using Corollary 2.14 in \cite{10.1561/2200000018}, EG achieves expected static regret guarantee
\begin{equation}
     \sum\nolimits_{t\in[T]}  \mathbb{E} f_t(\by_t;\theta_t)-\mathbb{E} f_t(\by^*;\theta_t) = O\left(\max\{(\bar{N}+B)/B, B\}\sqrt{T\log \bar{N}}\right),
\end{equation}
where $\by^* = \argmin_{\by\in \mathcal{Y}} \sum_{t\in[T]} \mathbb{E} f_t(\by;\theta_t)$ is the optimal decision.

Next, we show that we can compete against the offline optimal sequence of decisions with respect to the $\drcr$ $U_t(\by_t;\theta_t)$, which upper bounds $\mathbb{E} f_t(\by_t;\theta_t)$. Consider the formulation of the $\drcr$ in the problem~\eqref{p:ski-rental}: $U_t(\by_t;\theta_t)=(1-\delta)\eta_t + \delta \gamma_t$, where $\eta_t$ and $\gamma_t$ are chosen to be as small as possible while satisfying the constraints given $\by_t$. Here, we consider Lipschitzness of $U_t$ with respect to only $\delta$, which is sufficient due to the specific manner in which the $\epsilon$-net is constructed in \Cref{alg:col-app} (i.e., the $(\ell,u)$ coordinates of the points in the net are selected from a discrete set). Since $\eta_t$ and $\gamma_t$ are bounded by $\max\{(\bar{N}+B)/B, B\}$, $U_t$ is $\max\{(\bar{N}+B)/B, B\}$-Lipschitz in $\delta$. For brevity, we let $L := \max\{(\bar{N}+B)/B, B\}$. We also note that in Algorithm~\ref{alg:col-app}, any $\theta_t$ belonging to the same point of any $\mathcal{N}_{(i,j)}$ is within $\epsilon$ distance of each other. Thus, we can use the same steps of the proof of Theorem~\ref{thm:lipschitz cost}, which gives us
\begin{equation*}
    \sum\nolimits_{t \in \mathcal{T}_\theta}[\mathbb{E}f_t(\by_t;\theta_t)-{U_t(\by_t^*;\theta_t)]} \leq \Tilde{O}(L\sqrt{T_\theta}) + 2T_\theta \cdot \epsilon L,
\end{equation*}
where $\by_t = \pi_t(\theta_t)$ is the online decision by Algorithm~\ref{alg:col-app} and $\by^*_t = \pi^*(\theta_t)$ is the optimal decision for $U_t$.
Then, using Cauchy-Schwarz inequality,
\begin{align*}
    \sum\nolimits_{\theta\in \mathcal{N}} \Tilde{O}(L\sqrt{T_\theta}) + 2T_\theta \epsilon L &\leq \Tilde{O}(L\sqrt{|\mathcal{N}|T}) + 2T\epsilon L\\
    &= \Tilde{O}(L\sqrt{\sum\nolimits_{(\ell,u)}|\mathcal{N}_{(\ell,u)}|T}) + 2T\epsilon L\\
    &\leq \Tilde{O}(\bar{N} L\sqrt{T/\epsilon}) + 2T\epsilon L,
\end{align*}
where the last inequality follows because $|\mathcal{N}_{(\ell,u)}|=O(1/\epsilon)$. Setting $\epsilon=T^{-1/3}<1$ yields the final result.

\subsection{Proof of Corollary~\ref{cor:os-ol}}
\label{app:os-ol-proof}
Define two sets of discretized points $\Lambda_1=\{m(1+\lambda_1)^k\}_{k=0,\dots,K}$, where $K=\lfloor\frac{\ln(M/m)}{\ln(1+\lambda_1)}\rfloor$, and $\Lambda_2=\{m,m+\lambda_2,m+2\lambda_2,\dots,M\}$. Then we consider the discretization $\Lambda=\Lambda_1 \cup \Lambda_2$. Note that $\Lambda$ takes points from the discrete approximation of problem~\eqref{p:online-search} and the evenly spaced points in $[m,M]$. In the algorithm, for each $t \in [T]$, we round $(\ell_t,u_t)$ into points $(\Tilde{\ell}_t,\Tilde{u}_t)$ in $\Lambda_2 \times \Lambda_2$, where $\ell_t$ is rounded down and $u_t$ is rounded up. Then, Algorithm~\ref{alg:col-app} with EG as the master algorithm is run on $\Lambda$ and $(\Tilde{\ell}_t,\Tilde{u}_t,\delta_t)$. Here, EG runs over the simplex $\hat{\mathcal{Q}}$ supported on $\Lambda$ and optimizes the profit ratio function $f_t(\hat{\bq};\theta_t)$. Since $f_t$ is bounded above by $M/m$, EG with step size $\frac{\sqrt{\log (\frac{M-m}{\lambda_2}+K+1)}}{(M/m)\sqrt{2T}}$ yields the following static regret guarantee with respect to any $\hat{\bq}^* \in \hat{\mathcal{Q}}$:
\begin{align*}
    \sum\nolimits_{t\in[T]} [f_t(\hat{\bq}_t;\theta_t)-f_t(\hat{\bq}^*;\theta_t)] &= O\left((M/m)\sqrt{T\log (\frac{M-m}{\lambda_2}+K+1)}\right)\\
    &\leq O\left((M/m)\sqrt{T\log (\frac{M-m}{\lambda_2}+\ln(M/m)(1+\frac{1}{\lambda_1})+1)}\right),
\end{align*}
where the last inequality follows because $\ln(1+\lambda_1) \geq \lambda_1 / (1+\lambda_1)$ for $\lambda_1 > 0$. Ultimately, we will optimize the $\drcr$ over points in $\hat{\mathcal{Q}}$ and a rounded $(\Tilde{\ell},\Tilde{u},\delta)$. We claim that this still allows us to compete against the optimal point in ${\mathcal{Q}}$. To show this, we will need to (i) verify that the optimal $\drcr$ with respect to $\hat{\mathcal{Q}}$ is not far from the optimal $\drcr$ with respect to original feasible set $\mathcal{Q}$, and (ii) verify that the optimal $\drcr$ with $(\Tilde{\ell},\Tilde{u})$ is not far from the $\drcr$ with $(\ell,u)$, both being with respect to $\mathcal{Q}$.

For the first part, set $U_t(\bq;\theta_t)=(1-\delta_t)\eta_t + \delta_t \gamma_t$ as in the $\drcr$ formulation in problem~\eqref{p:online-search}.
Since $\eta_t,\gamma_t$ are bounded by $M/m$, $U_t$ is $M/m$-Lipschitz in $\delta$.
We will also consider another $\drcr$ upper bound $U'_t$ optimizing over $\hat{\mathcal{Q}}$ corresponding to problem~\eqref{p:online-search-d}, which is a relaxation of problem~\eqref{p:online-search}. For $\bq^* = \argmin_{\bq \in \mathcal{Q}}U_t(\bq;\theta_t)$ and $\hat{\bq}^* = \argmin_{\hat{\bq} \in \hat{\mathcal{Q}}} U'_t(\hat{\bq};\theta_t)$, Lemma~\ref{lem:os-discrete} yields $U'_t(\hat{\bq}^*;\theta_t)\leq U_t(\bq^*;\theta_t) + \frac{M}{m}\lambda_1$. This holds even though we added $\{m+\lambda_2,m+2\lambda_2,...,M\}$ to the original discretization in problem~\eqref{p:online-search-d} because adding more points to the discretization will only decrease the approximation error of the discrete $\drcr$. Thus, competing against $U'_t$ allows us to compete against $U_t$.

For the second part, note that the error incurred by rounding $(\ell, u)$ to $(\Tilde{\ell}, \Tilde{u})$ only affects $\eta$: in problem~\eqref{p:online-search}, constraints corresponding to $V\in [\Tilde{\ell},\ell) \cup (u,\Tilde{u}]$ will be added to constraint set~\eqref{eq:constr1} and removed from constraint set~\eqref{eq:constr2}. Thus, $\eta$ will only increase, while $\gamma$ is bounded below by $\eta$ and can only increase by the amount that $\eta$ increases since~\eqref{eq:constr2} has no added constraints. Focusing on constraints~\eqref{eq:constr1}, we consider how much $\eta$ can increase by adding constraints corresponding to $V=\Tilde{\ell}$ and $V=\Tilde{u}$. For any fixed $\bq$, denote
\begin{equation*}
    G(c)=\int_m^c v\cdot q(v) dv+(1-\int_m^c q(v) dv)m.
\end{equation*}
For the former, by adding the constraint
\begin{equation*}
    \Tilde{\ell} \leq \eta G(\Tilde{\ell}),
\end{equation*}
$\eta$ will change by at most
\begin{align*} 
    |\ell / G(\ell) -\Tilde{\ell} / G(\Tilde{\ell})| &\leq |\ell / G(\ell) -\ell / G(\Tilde{\ell})| + \lambda_2 / G(\Tilde{\ell})\\
    &\leq \ell\frac{G(\ell)-G(\Tilde{\ell})}{G(\ell)G(\Tilde{\ell})}+\lambda_2 / G(\Tilde{\ell})\\
    &\leq \ell \frac{G(\ell) - G(\ell)+\int_{\Tilde{\ell}}^\ell v\cdot q(v) dv-m\int_{\Tilde{\ell}}^\ell q(v) dv}{m^2} + \lambda_2 / m\\
    &\leq \ell (\ell^2-\Tilde{\ell}^2) / 2m^2 + \lambda_2 / m \leq \lambda_2(M^2 / m^2+1/m),
\end{align*}
where the first inequality follows by triangle inequality and the third inequality follows by taking $G(c)\geq m$ and $x(v)\leq 1$.
Similar steps for the latter yields the same bound:
\begin{align*} 
    |u / G(u) -\Tilde{u} / G(\Tilde{u})| &\leq |u / G(u) -u / G(\Tilde{u})| + \lambda_2 / G(\Tilde{u})\\
    &\leq u\frac{G(\Tilde{u})-G(u)}{G(u)G(\Tilde{u})}+\lambda_2 / G(\Tilde{u})\\
    &\leq u \frac{G(\Tilde{u}) - G(\Tilde{u})+\int_{u}^{\Tilde{u}} v\cdot q(v) dv-m\int_{u}^{\Tilde{u}} q(v) dv}{m^2} + \lambda_2 / m\\
    &\leq u (\Tilde{u}^2-u^2) / 2m^2 + \lambda_2 / m \leq \lambda_2(M^2 / m^2+1/m),
\end{align*}
Denote $\Tilde{\theta}_t=(\Tilde{\ell}_t,\Tilde{u}_t,\delta_t)$, and let $\Tilde{\bq}^*,\bq^*$ be the optimal decisions for $U_t(\cdot;\Tilde{\theta}_t),U_t(\cdot;\theta_t)$, respectively. Thus, $U_t(\Tilde{\bq}^*;\Tilde{\theta}_t) \leq U_t(\bq^*;\theta_t) + (\frac{M^2}{m^2}+\frac{1}{m})\lambda_2$. Finally, note that the number of points in the $\epsilon$-net constructed by \Cref{alg:col-app} is $|\mathcal{N}|=\sum_{(\ell,u)}|\mathcal{N}_{(\ell,u)}|=\mathcal{O}((\frac{M-m}{\lambda_2}+1)^2/\epsilon)$, and that $U'_t$ is $M/m$-Lipschitz in $\delta$. Putting everything together and again following the steps of the proof for Theorem~\ref{thm:lipschitz cost},
\begin{align*}
    \mathbb{E}\sum_{t \in [T]}f_t(\hat{\bq}_t;\theta_t) &\leq \Tilde{O}((\frac{M-m}{\lambda_2}+1)(M/m)\sqrt{T/\epsilon}) + 2T\epsilon (M/m) + \sum_{t\in [T]}U'_t(\hat{\bq}^*_t;\Tilde{\theta}_t)\\
    &\leq \Tilde{O}((\frac{M-m}{\lambda_2}+1)(M/m)\sqrt{T/\epsilon}) + 2T\epsilon (M/m) + T\lambda_1 (M/m) + \sum_{t\in [T]}U_t(\Tilde{\bq}^*_t;\Tilde{\theta}_t)\\
    &\leq \Tilde{O}((\frac{M-m}{\lambda_2}+1)(M/m) \sqrt{T/\epsilon} + (M^2/m^2+M/m)(\epsilon+\lambda_1+\lambda_2) T) + \sum_{t\in [T]} U_t(\bq^*_t;\theta_t).
\end{align*}
Here $\Tilde{O}$ hides factors of $\ln(M/m)$. Setting $\epsilon=\min\{(M/m)^{-2/5},1\}T^{-1/5}<1$, $\lambda_1=\min\{(M/m)^{-2/5},1\}\cdot\min\{M/m-1,1\}T^{-1/5} < M/m-1$, and $\lambda_2=\min\{(M/m)^{-2/5},1\}(M-m)T^{-1/5}<M-m$ yields the result.

\subsection{Experiments}
\label{app:experiment}
\paragraph{Setup.}

We set buying cost to $B = 2$.
We generate $T = 3000$ instances, each with true skiing days $n_t$ sampled uniformly at random from $\{1,\ldots,8\}$. Synthetic \pip predictions are generated by sampling a point $p_t$ from a normal distribution $\mathcal{N}(n_t,\sigma_t^2)$ and then taking the $90\%$ confidence interval $(\ell_t,u_t)=(p_t-z_{0.95}\sigma_t,p_t+z_{0.95}\sigma_t)$. Here, $\sigma_t$ is set to simulate oscillating good and bad predictors: the first $10$ instances have $\sigma_t=0$, followed by the next $10$ with $\sigma_t=6$, and repeating.

\paragraph{Comparison algorithms.} We run the following online algorithms over the same sequence of UQs and instances $10$ times and evaluate the average excess CR (i.e., the empirical ratio {minus} $1$) of these algorithms over these runs. \textbf{\texttt{WOA}}: worst-case optimal randomized algorithm~\cite{karlin1990competitive} that is $\nicefrac{e}{e-1}$-competitive.
\textbf{\texttt{FTP}}: follow-the-prediction algorithm that fully trusts the prediction; it buys immediately on day 1 if the prediction is less than $B$, and otherwise rents forever.
\textbf{\texttt{OL-Dynamic}}: online learning with respect to policy regret (Algorithm~\ref{alg:col} adapted to ski rental, i.e., Algorithm~\ref{alg:col-app}), i.e., competing against $\pi^* = \argmin_{\pi} \sum_{t\in[T]} U_t(\pi(\theta_t);\theta_t)$.
\textbf{\texttt{OL-Static}}: online learning with respect to static regret, i.e., competing against the optimal fixed decision 
without considering UQ predictions.
\textbf{\texttt{RSR-PIP}}: randomized algorithm with \pip (Algorithm~\ref{alg:os2}) that achieves the optimal \drcr.

\end{document}